\documentclass[review]{elsarticle}
\usepackage[whole]{bxcjkjatype}

\usepackage{bm}

\usepackage{amssymb}
\usepackage{amsmath}
\usepackage{color}

\usepackage[ruled,vlined]{algorithm2e}
\usepackage{algorithm2e,setspace}

\SetCommentSty{mycommfont}

\usepackage{booktabs}

\usepackage{graphicx}
\usepackage{floatrow}

\usepackage{array}
\newcolumntype{C}[1]{>{\centering\arraybackslash}p{#1}}

\usepackage{lineno,hyperref}
\usepackage{subfigure}

\newcommand{\tabref}[1]{Table~\ref{#1}}
\newcommand{\equref}[1]{Eq.~(\ref{#1})}
\newcommand{\figref}[1]{Fig.~\ref{#1}}
\newcommand{\chapref}[1]{Section~\ref{#1}}
\newcommand{\algref}[1]{Algorithm~\ref{#1}}

\newcommand{\bs}[1]{\boldsymbol{#1}}

\DeclareMathOperator*{\argmax}{argmax}

\makeatletter
\def\ps@pprintTitle{%
  \let\@oddhead\@empty
  \let\@evenhead\@empty
  \def\@oddfoot{\reset@font\hfil\thepage\hfil}
  \let\@evenfoot\@oddfoot
}
\makeatother

\begin{document}

\begin{frontmatter}

\title{
   Robust Iterative Value Conversion: Deep Reinforcement Learning for Neurochip-driven Edge Robots
}

\author[naistaddress]{Yuki Kadokawa\corref{cor1}}
\ead{kadokawa.yuki@naist.ac.jp}

\author[naistaddress]{Tomohito Kodera} \ead{kodera.tomohito.kp9@is.naist.jp}

\author[naistaddress]{Yoshihisa Tsurumine} \ead{tsurumine.yoshihisa@is.naist.jp}

\author[megaChips]{Shinya Nishimura} \ead{nishimura.shinya@megachips.co.jp}

\author[naistaddress]{Takamitsu Matsubara}
\ead{takam-m@is.naist.jp}

\cortext[cor1]{Corresponding author}

\address[naistaddress]{Nara Institute of Science and Technology, 630-0192, Nara, Japan}

\address[megaChips]{MegaChips Corporation, 532-0003, Osaka, Japan}

\begin{abstract}
A neurochip is a device that reproduces the signal processing mechanisms of brain neurons and calculates Spiking Neural Networks (SNNs) with low power consumption and at high speed. Thus, neurochips are attracting attention from edge robot applications, which suffer from limited battery capacity. This paper aims to achieve deep reinforcement learning (DRL) that acquires SNN policies suitable for neurochip implementation. Since DRL requires a complex function approximation, we focus on conversion techniques from Floating Point NN (FPNN) because it is one of the most feasible SNN techniques. However, DRL requires conversions to SNNs for every policy update to collect the learning samples for a DRL-learning cycle, which updates the FPNN policy and collects the SNN policy samples. Accumulative conversion errors can significantly degrade the performance of the SNN policies. We propose Robust Iterative Value Conversion (RIVC) as a DRL that incorporates conversion error reduction and robustness to conversion errors. To reduce them, FPNN is optimized with the same number of quantization bits as an SNN. The FPNN output is not significantly changed by quantization. To robustify the conversion error, an FPNN policy that is applied with quantization is updated to increase the gap between the probability of selecting the optimal action and other actions. This step prevents unexpected replacements of the policy's optimal actions. We verified RIVC's effectiveness on a neurochip-driven robot. The results showed that RIVC consumed 1/15 times less power and increased the calculation speed by five times more than an edge CPU (quad-core ARM Cortex-A72). The previous framework with no countermeasures against conversion errors failed to train the policies. Videos from our experiments are available: \url{https://youtu.be/Q5Z0-BvK1Tc}.
\end{abstract}

\begin{keyword}
\texttt neurochip \sep robot learning \sep deep reinforcement learning
\MSC[2020] 00-01\sep  99-00
\end{keyword}

\end{frontmatter}

% \linenumbers

\section{Introduction}

    Edge robots with limited battery capacity need to power-efficiently calculate control policies \cite{Robot_using_DRL_for_transport,transportation2, Robot_using_DRL_for_inspection,inspection2}.
    For that purpose, neurochips are attracting attention for implementing policies in power-efficient applications \cite{Learning_SNN_for_Robot, Robot_using_SNN,nc_snake_stdp}.
    A neurochip is a device that reproduces the signal transmissions and the processing mechanisms of brain neurons \cite{neurochip_hard_robot,akida1,loihi-dqn-quantize}.
    As in brains, a neurochip handles spike signals which are the binary representations of on/off signals.
    After spike signals are fired from multiple nodes and a certain threshold value is integrated, a specific spike signal is output; various outputs are expressed from spike-signal combinations \cite{neurochip_calc,akida2}.
    Neurochips can calculate functions encoded by spike-signal processing much faster and with less power consumption than such general-signal processing computers as CPUs and GPUs since neurochips have optimized memory and arithmetic units for spike-signal processing \cite{akida_hardware,akida1,loihi_hardware}.
    The function attracting the most attention as a neurochip application is the Spiking Neural Network (SNN) \cite{snn2,snn3}, which approximates the complex functions of high-dimensional inputs \cite{akida1,ibm-neurochip,loihi-dqn-quantize}.
    
\begin{figure}[t]
    \vspace{2mm}
    \centering
    \includegraphics[width=0.9\linewidth]{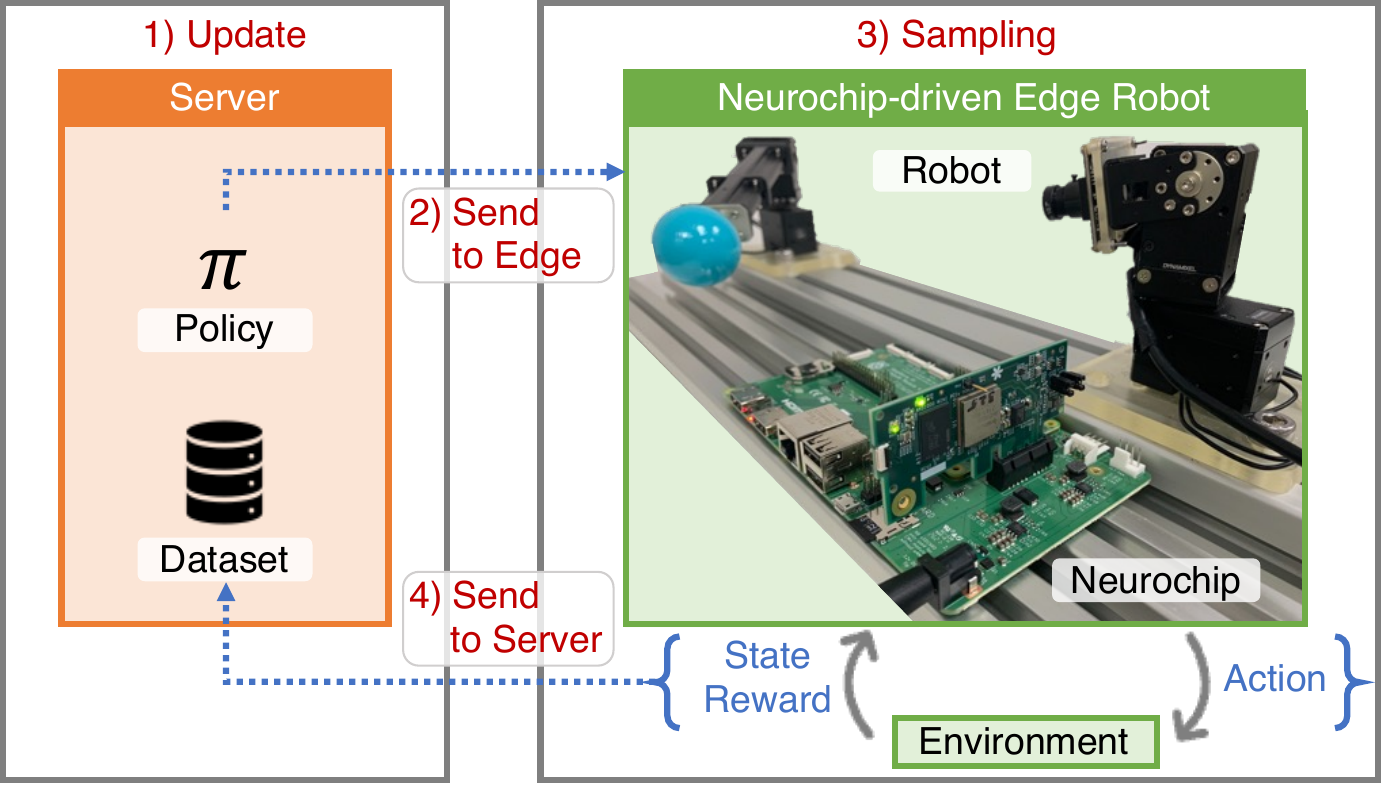}
    \caption{
        Learning scheme for neurochip-driven robot policies in proposed framework, which
        trains a policy of a neurochip-driven robot in real-world interactions. First, we create an edge-server-learning system that updates policies in the server and in the sample dataset in a neurochip-driven robot. 
        Learning flow: 
        1) server updates policies; 
        2) server sends them to neurochip; 
        3) neurochip-driven robot samples learning dataset; 
        4) neurochip-driven robot sends samples to server. 
        This cycle is conducted until policy converges.
    }
    \label{fig:overview}
\end{figure}

    This paper focuses on deep reinforcement learning (DRL) to obtain real-robot SNN policies suitable for neurochip implementation (\figref{fig:overview}).
    DRL automatically trains policies by mapping actions from complex high-dimensional observations through interaction with the environment.
    Almost every DRL calculated on CPU/GPU utilizes general Floating Point NN (FPNN) policies, which have been demonstrated in such fields as arcade games and robot control. A DRL with an FPNN successfully trained the complex policies of image input \cite{DRL_survey, dqn}.
    In recent years, techniques, which converted the trained FPNN policies obtained by such DRLs into SNN policies \cite{ANNtoSNN,ANNtoSNNviaPara}, have attracted attention because they obtain better total rewards than directly training SNN policies \cite{R-STDP, r-stdp-lane-keeping, loihi-dqn-quantize}. An SNN cannot utilize one of the most accurate DRL update schemes (called gradient descent) since spike signals are not directly differentiable. 
    This conversion consists of quantizing an FPNN so that the spike-firing frequency corresponds to the discrete value of the FPNN's quantized output and constructing an SNN from the quantized network \cite{akida1,loihi-dqn-quantize}.
    Previous works applied this conversion technique to a simulation task built on a CPU/GPU and utilized it as a policy conversion in a role that converted the trained FPNN policies into SNN policies \cite{ANNtoSNNforRL}.
    
    However, previous works have not been applied to real-robot DRLs, and the most significant factor is the conversion error caused by the policy conversion.
    A real-robot DRL is essential for collecting on-policy samples to reduce the number of training samples since real-robot samples suffer from high costs due to their slow operating speed and fragility to long operations \cite{DRL_survey}.
    To implement an on-policy DRL in a neurochip-driven robot, FPNN policies must be converted to SNN policies for every policy update to utilize the latter for on-policy sampling. However, such learning will fail due to accumulative conversion errors.
    Specifically, the first-stage NN quantization of a policy conversion introduces quantization errors, and the second-stage NN model conversion introduces model conversion errors, which replace the optimal actions of the SNN policy.
    
    To avoid the above problem, our approach trains FPNN policies to be robust against conversion errors, allowing accumulative conversion errors to be avoided due to repeated policy transformations at each policy update.
    We achieve this step by the following reductions and robustifications of the two-stage errors of policy conversion.
    To reduce the NN quantization error in the first stage, we utilize NN learning techniques that approximate the functions with a limited number of quantization bits, which have been developed in recent years in the field of image classification \cite{ANNtoSNNviaPara}. This approach prevents NN quantization from altering the actions of NN policies.
    As a second stage of robustness against NN model conversion errors, we seek optimal actions that do not change between the pre-converted FPNN policies applied with quantization and the post-converted SNN policies; in DRL, correctly obtaining optimal actions is sufficient \cite{cvi, Binarized_P-Network}.

    Bearing that two-stage idea in mind, we propose Robust Iterative Value Conversion (RIVC) as a DRL framework that is robust to the accumulative conversion errors resulting from policy conversion to obtain real-robot policies that can be implemented on a neurochip.
    RIVC trains policies by reducing and robustifying the errors in both stages of the policy conversion:
    1) RIVC directly trains quantized parameters to reduce quantization errors. We achieve this step by applying the learning rule of a quantized NN \cite{dorefa} to pre-converted NN policies.
    2) For robustness against the alternation of optimal actions due to model conversion errors, RIVC updates the policies to increase the gap between the probabilities of choosing optimal and non-optimal actions. We achieve this idea by applying an RL action-enhancement scheme called a gap-increasing operator \cite{gio_1,gio_2} to update the pre-converted NN policies.
    We evaluated the effectiveness of our proposed method on two simulation tasks and a real-robot visual-servo task. The proposed RIVC trained the SNN policies in the real world, although previous works without countermeasures for conversion errors cannot.
    Furthermore, SNN policies on neurochips consume 15 times less power and calculate five times faster than NN policies on edge CPUs (quad-core ARM Cortex-A72).
    
    The following are the contributions of this paper: 
    1) it proposed a novel DRL framework for training SNN policies in neurochip-driven robots; 2) it developed a new policy-update algorithm that suppresses the optimal-action changes caused by policy conversion;
    3) it verified energy savings and the acceleration of SNN policy calculation in a real-robot environment using neurochips.

    The rest of this paper is organized as follows.
    Section \ref{related-works} describes related works. 
    In Section \ref{preliminaries}, we offer preliminary considerations before discussing our proposed RIVC method, which is described in
    Section \ref{proposed-method}. 
    Section \ref{experimens} presents experiments composed of simulation trials (Section \ref{simulation-experimens}) and real-robot experiments (Section \ref{real-experiments}). 
    In Section \ref{discussion}, we discuss RIVC, and
    Section \ref{conclusion} concludes this paper.

\section{Related Works}
\label{related-works}

    \subsection{Definition of Learning Settings}
        This paper focuses on DRL for real robots, particularly battery-limited edge ones, which are significantly affected by power consumption and calculation speed.
        There are three frameworks for the training policies of edge robots.

        \textbf{Edge Learning:}
            An edge robot collects real-robot samples. The neurochips and edge CPUs are implemented in its update policies. Edge learning can acquire policies using just edge robots \cite{edge1}.
            However, neurochips and edge CPUs cannot quickly update policies since the former cannot calculate the BPs; the latter cannot quickly calculate the BPs due to limited calculation resources \cite{edge-server3}.

        \textbf{Server Learning:}
            The server collects samples in a simulation and updates the SNN policies. 
            Simulators can collect learning samples more deftly than real robots since they can easily and quickly run robots and in parallel \cite{snn-sim2real}.
            However, this method suffers from the tremendous engineering cost of designing a simulator that reproduces a real-robot phenomenon \cite{r-stdp-sim2real}.

        \textbf{Edge-server Learning:}
            An edge robot collects real-robot samples and sends them to a server, which updates the policies and returns the updated policies to the edge robot. These steps are repeated until the learning is completed \cite{Binarized_P-Network,edge-server1,edge-server2}. The GPU server conducts the policy updates to speed up the learning process since DRL's backpropagation calculations are time-consuming.

        This paper utilizes such an edge-server-learning style for two reasons:
        (1) Neurochips cannot calculate BPs.
        (2) The learning process is accelerated since edge learning requires tremendous waiting time to update policies.

\begin{table}[t]
        \caption{
           Comparison of proposed method and related works:
           ``Implementable in neurochips'' denotes methods that can implement SNN policies in neurochips.
           ``High-dimensional observation'' denotes methods that can train such policies, including image input.
           ``Learning framework'' denotes method categorization in three terms: 1) ``Edge'' means learning in just an edge environment. 2) ``Server'' means learning in just a server environment. 3) ``Edge-Server'' denotes learning in both server and edge environments (a sampling learning dataset in edge environment and updating policies in server).
           \label{table:previous_compare}
        }
        \small
        \begin{tabular}{lccc}
            \toprule
                &\textbf{Implementable} & \textbf{High-dimensional} & \textbf{Learning} \\
                &\textbf{in Neurochips} & \textbf{Observation} & \textbf{Framework} \\
            \midrule
                \textbf{R-STDP} \cite{R-STDP_using_robot,r-stdp-lane-keeping} & \checkmark & - & Edge \\
                \textbf{SNN-BP} \cite{snn-with-designed-parameters-1,snn-with-designed-parameters-2} & - & \checkmark  & Server \\
                \textbf{DRL2SNN} \cite{ANNtoSNNforRL,drl2snn} & \checkmark & - & Server \\
                \textbf{RIVC} (Ours) & \checkmark & \checkmark & Edge-Server \\ 
            \bottomrule
        \end{tabular}
\end{table}

    \subsection{Reinforcement Learning with Spiking Neural Networks}
        DRL frameworks utilizing SNNs are categorized as shown in \tabref{table:previous_compare}.
        This section describes the methods for training SNN policies and their applicability to real robots.

        \subsubsection{Reward-modulated Spike Timing Dependent Plasticity (R-STDP)}
            This method trains SNN policies by imitating the learning flow of the human-brain structure, which updates the brain based on spike output timing.
            STDP increases or decreases the weights between the coupled nodes based on the timing of the spikes. This method increases the weights when the spikes on the function's input side fire first and the output-side neurons fire later; it decreases the weights in reverse situations. The weights are updated more significantly when a big reward matches the timing of the firing of the spike signal of the output-layer node that represents each action \cite{R-STDP, R-STDP_using_robot, r-stdp-lane-keeping}.
            This method can be implemented in numerous neurochips because it is straightforward and orthodox with few variables composed of Leaky integrate-and-fire (LIF) or IF neurons \cite{loihi-dqn-quantize,ibm-neurochip}.
            However, it does not work with SNNs that have many nodes, including CNNs. This problem is caused by the fact that it cannot accurately approximate policies for subtle reward differences since it just changes the policies in response to a large or small reward.

        \subsubsection{Training SNN-policy with Backpropagation (SNN-BP)}
            This method updates SNN weights by BPs. 
            SNN weights cannot be directly updated using BPs since the spike signals (output 0 or 1) are undifferentiable.
            Thus, this method utilizes the spike output's frequency to estimate continuous values for calculating gradients.
            It utilizes additional internal variables and increases the accuracy of the parameters compared to the IF and LIF neurons so that the SNN policies can accurately output actions reflecting subtle differences in rewards \cite{DLinSNN,snn-with-designed-parameters-1,snn-with-designed-parameters-2}.
            This method can be applied to image-input DRLs because internal variables were added.
            However, such additional internal variables are often unsupported by neurochips since they are highly dependent on the designer's implementation.
            Therefore, applying this method to DRL with SNN policies is difficult in real robots because implementing SNN policies on neurochips faces some manufacturing challenges.

        \subsubsection{DRL-policy to SNN-policy Conversion (DRL2SNN)}
            This method acquires SNN policies by model conversion from trained FPNN policies in two steps:
            1) It first trains the FPNN policies with some DRL algorithms (such as Deep Q-Network \cite{DRL_survey}) in the CPUs or the GPUs.
            2) Then the trained FPNN policies are converted to SNN policies for neurochip implementation \cite{ANNtoSNNforRL, ANNtoSNN, ANNtoSNNviaPara}.
            The SNN policies obtained from this method can be implemented in numerous neurochips because an SNN policy is composed of IF or LIF neurons \cite{akida1,loihi-dqn-quantize,ibm-neurochip}.
            This method can be applied to complex image-input tasks, such as ATARI games due to FPNN's high-function-approximation accuracy \cite{ANNtoSNNforRL}.
            
            Unfortunately, this method is not applicable to real-robot learning because it does not address conversion errors.
            Simulation learning trains only the FPNN policies and converts them to SNN policies just once at the end of learning since robot environments are built in the CPUs/GPUs.
            Learning on neurochip-driven robots needs to collect learning samples from the robot itself.
            In this case, we must iteratively convert the FPNN policies updated on the CPU/GPU to SNN policies to collect samples by SNN policies on the neurochip-driven robot.
            As a result, real-robot learning fails due to accumulative conversion errors because this method is not robust to them.
            
            In another naive approach, FPNN policies trained in simulations can be converted to SNN policies for real robots.
            However, this approach is undesirable because it raises additional concerns about reality gaps and the robot's communication system.
            Therefore, this paper focuses on training policies for neurochip-driven robots from learning samples from neurochip-driven robots.

        \subsubsection{Comparison with Our Method}
            In summary, the previous methods face certain challenges: 1) R-STDP cannot be applied to image-input tasks due to the limitations of function approximation accuracy; 2) SNN-BP cannot be applied to neurochip implementation; 3) SNN-FT requires a simulation environment; and 4) DRL2SNN cannot learn in real robots.
            From these problems, image-observation DRLs on real robots controlled by neurochips do not exist.
            
            To obtain complex SNN policies (for neurochip-driven robots) learned in the real world, we developed a novel DRL method inspired by DRL2SNN.
            We expanded DRL2SNN applicability to learning on neurochip-driven robots. It repeatedly converted from NNs to SNNs for each NN update for sampling datasets by SNN policies.
            However, our experiments show that this repeated conversion complicates learning because conversion errors are accumulated.
            We solve this problem by proposing a novel DRL method that reduces the effect of cumulative conversion errors.

    \subsection{Robotic Applications of Spiking Neural Networks and Neurochips}
        Robot control utilizing SNNs and neurochips has been studied due to its low power consumption and fast calculation speed. In this section, we summarize the application of SNN policies to neurochip-driven robots, focusing on their input sensors, and discuss the differences from our work.
        
        \subsubsection{Dynamic Vision Sensor}
            Previous research used a Dynamic Vision Sensor (DVS) to perform real-time visual information processing. Real-time tasks that have applied DVS include a target-tracking task using a wheel-less snake-like robot \cite{nc_snake_stdp} and a manipulator that tracks a moving ping-pong ball in real-time \cite{nc_pinpon_handtune_noLearning}. These SNN policies use event data from DVS as input and robot joint angles as output. 
            Another example is the use of DVS for high-speed control of an Unmanned Aerial Vehicle (UAV), which requires a limited battery size due to its light weight for flight \cite{nc_drone_pid}. Therefore, the combination of DVS and a neurochip is effective in real-time UAV control because it enables visual information processing at high speed and low power consumption \cite{nc_drone_generationAlg}.
            
            These studies use a DVS to input real-time visual information to an SNN for high-speed robot motion control and object tracking. However, since a DVS observes luminance change, it is unable to recognize stationary objects and complex-shaped objects.

        \subsubsection{Low-Dimensional Sensors}
            Previous research contains several examples of robot applications using low-dimensional sensor information. For example, a sound sensor was utilized for voice-based robot behavior, and an SNN was used for a voice discrimination function \cite{nc_low_dim_sensor}. Another example, although not directly used for control, exploits temperature sensor information to train an SNN model that switches the robot's behavior \cite{nc_low_dim_sensor}. Another study used an SNN to approximate the function of a central pattern generator, which is a walking pattern generator, to obtain an SNN model that switches the walking pattern of a six-legged robot based on signals given in real-time \cite{nc_6leg_cpg}.
            
            In these studies, low-latency robot control is achieved by converting sensor information into spike signals. However, these sensors are unsuitable for complex environment recognition compared to visual information.

        \subsubsection{Frame-Based Camera}
            Compared to previous research, this study uses a frame-based camera to perform visual information processing in complex environments. A frame-based camera can represent the environment with multiple pixels, enabling more detailed recognition than a DVS, which only captures luminance changes.

            Although there are several task applications of frame-based cameras in previous research, they have not been applied to a real-robot RL. For example, an autonomous wheeled robot on a mountain trail has been achieved \cite{nc_mobile_imitationLearning_cnn,nc_low_dim_sensor}; control policies based on frame-based image input have been obtained. In this research, FPNN policies were learned from a dataset collected by a human driver through controller commands, and the FPNN policies were converted to SNNs \cite{nc_mobile_imitationLearning_cnn}.

            Frame-based cameras, which enable the detection and tracking of stationary objects, are particularly useful for tasks that require detailed environmental recognition. Another advantage of frame-based cameras is that they are compatible with many existing computer vision algorithms and deep learning models, allowing the use of existing technology and software without significant modification. Because of these advantages, this study focuses on Reinforcement learning (RL), which is an SNN control policy using image input from frame-based cameras.

\section{Preliminaries} 
\label{preliminaries}

    \subsection{Reinforcement Learning}
        \label{sec:rl}
        Reinforcement learning, which optimizes an agent's actions in an environmental model that mirrors the Markov Decision Process (MDP),
        is comprised of the following five components: $\mathcal{S}, \mathcal{A}, \mathcal{T}, r, and \gamma$.
        $\mathcal{S}$ is a set of observations that can be obtained from the environment, and $\mathcal{A}$ is a set of selectable actions.
        $\mathcal{T}^a_{s s'}$ is the probability of transitioning to observation $s' \in \mathcal{S}$ when action $a \in  \mathcal{A}$ is chosen in observation $s \in \mathcal{S}$. 
        The reward for making the transition is represented by $r^a_{s s'}$, and $\gamma \in [0,1)$ is a discount factor.
        Policy $\pi(a|s)$ is the probability of choosing action $a$ in the case of observation $s$.
        The goal of RL is to find optimal policy $\pi^{*}$ that maximizes discounted total reward $\sum_{\substack{t = 0}}^{\infty} \gamma^{t} r_{s_{t}}$, where $r_{s_{t}} \!=\! \sum_{\substack{a_t \in \mathcal{A} \\ s_{t+1} \in \mathcal{S}}}\pi(a_t|s_t)\mathcal{P}^{a_t}_{s_t s_{t+1}}r^{a_t}_{s_t s_{t+1}}$.
        For each observation $s$, state value function $V^{\pi}$ under policy $\pi$ can be defined:
        \begin{eqnarray}
            \label{V_function}
            \begin{aligned}
                V^{\pi}(s)={\mathbb{E}}_{\pi, T}\bigg[\sum_{\substack{t = 0}}^{\infty} \gamma^{t} r_{s_{t}} \bigg| s_{0}= s \bigg].
            \end{aligned}
        \end{eqnarray}
        The RL objective is to find optimal policy $\pi^{*}$ that satisfies the Bellman equation:
        \begin{eqnarray}
            \label{V_Bellman}
            \begin{aligned}
                V^{*}(s) = \displaystyle\max_{\pi} \sum_{\substack{a \in \mathcal{A} \\ s' \in \mathcal{S}}} \pi(a|s) \mathcal{T}_{ss'}^{a} \big(r_{ss'}^{a} + \gamma V^{*}(s')\big), 
            \end{aligned}
        \end{eqnarray}
        where $V^{*}$ is the optimal state value function.
        To evaluate the policies based not only on observations $s$ but also actions $a$, the optimal action-value function is defined:
        \begin{eqnarray}
            \label{Q_Bellman}
            \begin{aligned}
                Q^{*}(s, a) \hspace{-0.05cm}=\hspace{-0.05cm} \displaystyle\max_{\pi} \hspace{-0.05cm} \sum_{s' \in \mathcal{S}}\mathcal{T}_{ss'}^{a}\big(r_{ss'}^{a} \hspace{-0.05cm}+\hspace{-0.05cm}  \gamma \hspace{-0.1cm}  \sum_{a' \in \mathcal{A}} \hspace{-0.05cm} \pi(a'|s')Q^{*}(s', a')\big), \hspace{-0.25cm}
            \end{aligned}
        \end{eqnarray}
        where $Q^{*}$ is an optimal Q function.

    \subsection{Gap-Increasing Operator}
        A Gap-Increasing Operator (GIO) is a RL technique for robustly updating value functions to function approximation errors \cite{gio_1,gio_2,cvi}.
        The Bellman equation is modified for using GIO:
        \begin{eqnarray}
            \label{eq:CVI-objective}
            V^{*}(s) = \displaystyle\max_{\pi} \! \sum_{\substack{a \in \mathcal{A} \\ s' \in \mathcal{S}}} \pi(a|s) \bigg[ \mathcal{T}_{ss'}^{a} \big(r_{ss'}^{a} \! + \! \gamma V^{*}(s')\big) \!+\! i_{\bar{\pi}}^{\pi}(s)\bigg], \!
        \end{eqnarray}
        where $i_{\bar{\pi}}^{\pi}$ is defined with current policy $\pi$ and baseline policy $\bar{\pi}$:
        \begin{eqnarray}
            \label{eq:CVI-objective-Constraint}
            i_{\bar{\pi}}^{\pi}(s) \!=\! \sum_{a \in \mathcal{A}} \! \pi(a|s) \bigg[\! -\frac{1\!-\!\alpha}{\beta} \log \pi(a|s) \! - \frac{\alpha}{\beta} \log \frac{\pi(a|s)}{\bar{\pi}(a|s)} \bigg], \!
        \end{eqnarray}
        where $\alpha \in [0,1]$ and $\beta \in (0,\infty)$ are hyperparameters.
        In contrast to the Q function, the action preference function, denoted by $P$, is defined:
        \begin{equation}
            \begin{split}
                \label{eq:CVI-action-value-function} 
                P^{\pi}(s, a) =& \sum_{\substack{s' \in \mathcal{S}}} \mathcal{T}_{ss'}^{a} (r_{ss'}^{a} + \gamma \sum_{a \in \mathcal{A}} \pi(a|s) V^{\pi}(s', a')) \\
                & + \frac{\alpha}{\beta}\log \pi(a|s). 
            \end{split}
        \end{equation}
        To find optimal policy $\pi^*$ that maximizes \equref{eq:CVI-action-value-function}, the update rule of action preference $P$ is defined:
        \begin{equation}
            \begin{split}
                \label{eq:CVI-update}
                P_{k+1}(s, a) \gets & r_{ss'}^{a}+ \gamma (m_{\beta} P_{k})(s') + \mathcal{G}(s,a), \\
                \mathcal{G}(s,a) & = \alpha \bigg(P_{k}(s, a) - (m_{\beta} P_{k})(s)\bigg), 
            \end{split}
        \end{equation}
        \vspace{-1mm}
        \begin{equation}
            \label{eq:mellowmax}
             \left( m_{\beta} P \right) (s) = \frac{1}{\beta} \log \left( \frac{1}{|\mathcal{A}|} \sum_{a \in A}\exp \left(\beta P(s,a)\right) \right), 
        \end{equation}
        where $|\mathcal{A}|$ is the number of selectable actions.
        The policy is given as follows: 
        \begin{equation}
            \label{eq:policy}
            \pi_{k}(s,a) = \frac{\exp \left(\beta P_{k}(s,a)\right)}{\sum_{b\in A} \exp \left(\beta P_{k}(s,b)\right)}. 
        \end{equation}
        $\mathcal{G}(s,a)$ in \equref{eq:CVI-update} is a GIO that amplifies the differences between the maximum value and others. Therefore, it makes the resulting policy for choosing optimal action robust against function approximation errors \cite{cvi}. 
        We refer to $\alpha$ as the GIO coefficient.
        When it is high, the robustness of the function approximation errors increases.
       $\beta$ controls the learning convergence.
        When it is high, the learning convergence becomes faster.

    \subsection{Quantized Neural Network}
        A Quantized Neural Network (QNN) is an NN calculated in low-bit quantization. Its weights and activation function outputs are quantized at a lower bit than traditional 32-bits or 16-bits \cite{dorefa}.
        The following sections describe QNN's structure and how to update its parameters.

        \subsubsection{Network Structure}
        
            Assuming that the dimensions of the input and output vectors in each NN layer are $N$ and $M$, each layer consists of a Fully-Connected Layer (FCL) and an activation function.
            NN's network parameters of the $l$-th layer (a completely $L$ layer) are $\bs{W}_l =[\bs{W}_{l,1}\bs{W}_{l,2}\dots \bs{W}_{l,M}] \in  \mathbb{R}^{N \times M}$, $\bs{W}_{l,m} =[w_{l,m,1} \ w_{l,m,2} \dots w_{l,m,N}] \in \mathbb{R}^{N}$ set to $\bs{\theta}=\{\bs{W}_1, \bs{W}_2, \dots, \bs{W}_L \}$.
            Let $x_l$, $\mathcal{F}$, $\mathcal{Q}$ denote each layer's outputs, activation, and quantization functions.
            $L$-layer QNN output ${x}_{l}$ in each layer is calculated as
            \begin{align}
                \label{eq:fnn-layer}
                {x}_{l,m} = \mathcal{F} \left( \sum_{n=0}^{N} \mathcal{Q}({w_{l,m,n}^{f}) x_{l-1,n}} \right), 
            \end{align}
            where $m$ and $n$ are the output and input node numbers. 
            $x_L$ denotes the QNN function approximation result.
            Activation function $\mathcal{F}$ and quantization function $\mathcal{Q}$ are
            \begin{align}
                \label{eq:quantize_activate_func}
                \mathcal{F}_k(x) = \frac{1}{2^k - 1}{\rm round} \left( \left( 2^k - 1 \right) x \right),
            \end{align}
            \begin{align}
                \label{eq:quantize_weight_func}
                \mathcal{Q}(w_{l,m,n}) = 2 \; \mathcal{F}_k \left( \frac{\tanh(w_{l,m,n})}{2 \underset{w_{l,i,j} \in W_{l}}{\max}(|\tanh(w_{l,i,j})|)} + \frac{1}{2} \right) - \frac{1}{2},
            \end{align}
            where $k$ is the number of quantization bits defined as a multiple of two and ${\rm round}$ is a quantization function that rounds the input to an integer.
            In computers, a policy can implement $\bs{\theta}^{q}$ with $k$-bits, reducing the model size to about $k$/32 compared to $\bs{\theta}^{f}$. Thus, QNNs reduce the calculation speed more than FPNNs \cite{qnn_lowenegy}.

        \subsubsection{Learning Parameters}
            QNN parameters are updated by 32-bit weights $\bs{\theta}^{f}$ since quantized weights $\bs{\theta}^{q}$ are insufficiently accurate to approximate the gradients for backpropagation.
            $\bs{\theta}^{f}$ is updated by loss function $\mathcal{L}_{f}$:
            \begin{equation}
                \label{eq:qnn-error} 
                \mathcal{L}_{f}(x; \bs{\theta}^q) = \frac{1}{2} \left[ Y(x;{\bs{\theta}^{q}}) - y  \right]^2, 
            \end{equation}
            where $Y$ approximates the QNN function and $y$ is the label.
            $\bs{\theta}^{f}$ are sequentially quantized to $\bs{\theta}^{q}$ for estimating the loss of \equref{eq:qnn-error}.
            The number of quantization bits can be arbitrarily determined by the weights and the activation level.
            The quantization of the weights follows \equref{eq:quantize_weight_func}, and the quantization of the output value of each node follows \equref{eq:quantize_activate_func}.
            In addition, only when $\bs{\theta}^{f}$ is updated, the weights are
            \begin{align}
                \label{eq:quantize_weight_func_qpn}
                \mathcal{Q}(w_{l,m,n}) = 2 \; \mathcal{F}_k \left( \frac{\tanh(w_{l,m,n})}{2 \underset{w_{l,i,j} \in W_{l}}{\max}(|\tanh(w_{l,i,j})|)} + \frac{1}{2} + \mathcal{N}(k) \right) - \frac{1}{2},
            \end{align}
            where $\mathcal{N}(k)$ is a noise function:
            \begin{align}
                \label{eq:quantize_weight_noize}
                \mathcal{N}(k) = \frac{{\rm Uniform} (-0.5,0.5)}{2^k - 1}.
            \end{align}
            Since $\mathcal{N}$ improves the accuracy of identification tasks \cite{dorefa,imagenet}, we improved the function approximation accuracy with it.

    \subsection{Converting FPNN to SNN}
        \label{howtoFPNN2SNN}

        An FPNN can be converted approximately to an SNN of the integrate-and-fire (IF) model.
        Each SNN layer is calculated:
        \begin{align}
            \label{eq:snn_node}
            x_{l,m}(t) = \sum_{n=0}^N w^{s}_{l,m,n} \mathcal{F}_s\big(x_{l-1,n}(t)\big) \\
            \label{eq:snn_activation}
            \mathcal{F}_s(x) = \left \{
                    \begin{array}{l}
                        1, \;\;\;\;\;\; \text{if} \ x > \mathcal{T}, \\
                        0, \;\;\;\;\;\; {\text{otherwise}}, 
                    \end{array}
                \right.
        \end{align}
        where $\mathcal{F}_s$ is an activation function, $\mathcal{T}$ is a firing threshold, and the SNN output is $y^s = \sum_{t} x_L(t)$.
        The $x$ values are reset to $0$ when $\mathcal{F}_s(x)$ outputs 1.
        
        An FPNN can be converted approximately to an IF model's SNN when 1) the bias term is zero and 2) the following ReLU activation function is used for the output results of each layer \cite{ann2snn_origin}:
        \begin{align}
            \label{eq:ReLU}
            {\rm ReLU}(x) = \max(0,x).
        \end{align}
        Following the above conditions, FPNN weight $\bs{\theta}^{f}$ can be converted to SNN weight $\bs{\theta}^{s}$ by removing the parameter biases from all the layers of $\bs{\theta}^{f}$ \cite{ann2snn_origin}.
        The remaining conversion manner (unrelated to this paper) is described in these works \cite{ANNtoSNN, ANNtoSNNviaPara, ANNtoSNNforRL}.
        
        SNN's function approximation accuracy is considerably lower than that of the FPNN since it was calculated from the sum of a finite number of spike signals.
        Conversion from FPNN to SNN decreases the function approximation accuracy due to conversion errors.
        Our proposed learning framework is robust to such function approximation error since it is fatal to RL, which requires function approximation accuracy.
        This paper develops a novel RL method that updates the QNN policies instead of the FPNN policies to reduce the conversion errors. Our learning framework is also robust to SNN conversion because it increases the values between the maximum action and other actions.

\section{Robust Iterative Value Conversion}
\label{proposed-method}

    This paper proposes Robust Iterative Value Conversion (RIVC), a novel DRL framework that enables the training of SNN policies by interaction between neurochip-driven robots and real-world environments.
    RIVC has two features: 1) it optimizes policies with quantized neural networks to avoid significant quantization conversion, and 2) it applies a gap-increasing operator to policy updates to emphasize the optimal actions for robustification against the unexpected replacements of policy actions for conversion errors that linger after applying the quantization errors.
    Our proposed RIVC is then applied to an edge-server-learning framework, which updates and converts policies using CPU/GPU and collects learning samples using neurochips.
    The details of the proposed method are described below, and the entire framework is summarized in \algref{alg:edge-server}.

\begin{algorithm}[t]
    \SetKwData{Left}{left}\SetKwData{This}{this}\SetKwData{Up}{up}
    \SetKwFunction{Union}{Union}\SetKwFunction{FindCompress}{FindCompress}
    \SetKwInOut{Input}{input}\SetKwInOut{Output}{output}
    \caption{Robust Iterative Value Conversion}
    \label{alg:edge-server}
    Set parameters described in \tabref{table:learning_setting} \\
    Set network weights ${\bs{\theta}^f}$, ${\bs{\theta}^q}$, $\bs{\theta}^{-q}$, ${\bs{\theta}^s}$ replay buffer ${\mathcal{D}}$ \\
    \SetKwFunction{DC}{\color{blue} Sampling Dataset}
    \SetKwFunction{UpPN}{\color{blue} Update Network}
    \SetKwFunction{Convert}{\color{blue} SNN-Policy Conversion}
    \SetKwProg{Fni}{\color{blue} i)}{}{\KwRet}
    \SetKwProg{Fnii}{\color{blue} ii)}{}{\KwRet}
    \SetKwProg{Fniii}{\color{blue} iii)}{}{\KwRet}

    \For{$ i = 1, 2, ..., I$}
    {   
        \Fni{\Convert}
        {
            $\bs{\theta}^{q} = $ Quantize $\bs{\theta}^{f}$ by \equref{eq:quantize_weight_func} \\
            $\bs{\theta}^{s} = \text{Convert} \ \bs{\theta}^{q}$ described in \chapref{howtoFPNN2SNN} \\
        }
        
        \Fnii{\DC}
        {
            Initialize replay buffer ${\mathcal{D}}$ \\
            \For{$ e = 1, 2, ..., E$}
            {
                \For{$ t = 1, 2, ..., T$}
                {
                    Take action $a_t$ with softmax policy \equref{eq:policy} based on ${P}(s_t, \mathcal{A};{\bs{\theta}}^{s})$ \\
                    Receive observation $s_{t\!+\!1}$, reward $r_{s_t s_{t\!+\!1}}^{a_t}$ \\
                    Push $\{(s_t, a_t,r_{s_t s_{t+1}}^{a_t},s_{t+1})\}$ to $\mathcal{D}$ \\
                }
            }
        }
    
        \Fniii{\UpPN}
        {
            Set target network $\bs{\theta}^{-f} = \bs{\theta}^f$ \\
            \For{$ c = 1, 2, ..., C$}
            {
            Set ${\mathcal{D}'}$ is index-shuffle local memory ${\mathcal{D}}$ \\
            \For{$ k = 1, 2, ..., \rm{round}(\ |\mathcal{D}| / B\ )$}
            {
              Sample the minibatch of transition ${\mathcal{D}'}[B \! \times \! (k\!-\!1) : B \! \times \! k]$ \\
              $\bs{\theta}^{q} = $ Quantize $\bs{\theta}^{f}$ with noise injection by \equref{eq:quantize_weight_func_qpn} \\
              Calculate loss on \equref{eq:qpn-error} by $\bs{\theta}^{q}$ then update $\bs{\theta}^f$ \\
              }
            }
        }
    }
\end{algorithm}

\begin{table}[]
    % \vspace{2mm}
    \caption{
            Learning parameters of proposed framework:
            Simulation and real-robot experiments generally follow parameters.
           Different parameters are described in each experiment section.
            \label{table:learning_setting}
    }
    \vspace{-2mm}
    \begin{center}
        \begin{tabular}{@{}lp{5.5cm}llll@{}}%{@{}lp{10cm}lp{10cm}l@{}}%
            \toprule
            \textbf{Parameter} & \textbf{Meaning} & \textbf{Value}  \\ 
            \midrule
            $B$ & Minibatch size & 32 \\
            $C$ & Number of epochs & 1000 \\
            $I$ & Number of iterations & 50 \\ 
            $E$ & Number of episodes per iteration & 10 \\
            $T$ & Number of steps per episode & 100 \\
            $U$ & Number of samples in $\mathcal{D}$ & $5 \times E \times T$ \\ 
            $\sigma$ & Maximum value of quantization & 1.00 \\
            $\alpha$ & Error robustness coefficient of GIO \cite{cvi} & 0.99 \\
            $\beta$ & Learning speed coefficient of GIO \cite{cvi} & 1.00 \\
            $\gamma$ & Discount factor of RL & 0.97 \\ 
            \bottomrule
        \end{tabular}
    \end{center}
\end{table}

\begin{figure}[]
    \centering
    \includegraphics[width=0.95\linewidth]{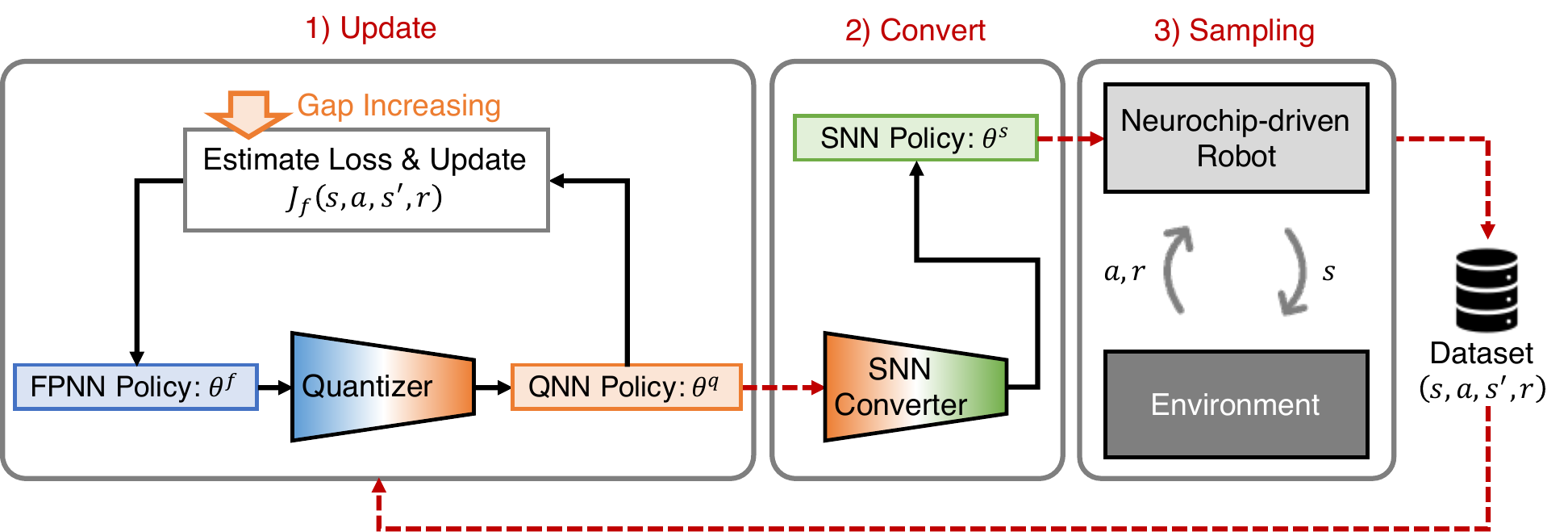}
    \caption{
        RIVC's learning framework:
        1) policy updates, 2) SNN conversion, and 3) a sampling dataset: 
        1) This step is proposed framework's main part. This update scheme trains QNN policies that prevent maximum action of policies from being replaced due to SNN conversion
       by increasing value gap of QNN policies between maximum action of policies and other actions.
        First step obtains a value of the QNN policy.
        Next update scheme estimates loss function by determining target value (including gap-increasing operator) to increase differences between estimated maximum action and other actions.
        FPNN policies are updated based on estimated loss function to more accurately calculate gradient with FPNN parameters with larger bits than QNN parameters.
        Updated FPNN parameters are quantized to QNN parameters, including noise injection into former to stabilize parameter updates.
        2) Trained QNN policy is converted to SNN policy.
        3) Neurochip-driven robots collect samples by SNN policy.
        Above three steps are repeated until policy converges.
    }
    \label{fig:learning_system}
\end{figure}

    \subsection{Learning Framework}
        Our proposed learning framework is shown in \figref{fig:learning_system}.
        The proposed method handles three types of policies and parameters: FPNN parameters $\bs{\theta}^{f}$ for calculating the high-accuracy gradients for the QNN update, QNN policy $\bs{\theta}^{q}$ for approximating the functions during the QNN updates, and SNN policy $\bs{\theta}^{s}$ for collecting the training samples by SNNs implemented in the neurochips.
        
        The learning process consists of the following steps:
        Convert $\bs{\theta}^{f}$ to $\bs{\theta}^{q}$ to $\bs{\theta}^{s}$. 
        The robot collects a dataset of state-action samples $a$, observations $s$, and rewards $r$ from the environment using SNN policy $\bs{\theta}^{s}$.
        Then this framework updates QNN policies $\bs{\theta}^{q}$ using the training dataset and
        repeatedly converts FPNN parameters $\bs{\theta}^{f}$ to SNN policies $\bs{\theta}^{s}$, which have some conversion errors.
        Thus, we propose RIVC, including a conversion-aware update scheme, which is robust to the conversion errors described in the following sections.

    \subsection{Reduction of Conversion Errors by Quantized Neural Network}
        Our proposed method trains policies not with a 32-bit FPNN but with quantized weights that reduce the conversion errors to SNN policies since converting FPNN to SNN produces significant quantization errors \cite{loihi-dqn-quantize}.
        These conversion errors sometimes replace optimal actions from FPNN to SNN.
        Inspired by these works, reducing the quantization-bit number allows us to retain the optimal actions from FPNN to SNN.
        A previous study also focused on the 1-bit case \cite{fnn2bnn-snn-1, fnn2bnn-snn-2, fnn2bnn-snn-3}, although it needs to clarify which bit number from 1- to 32-bits is the most suitable for the conversion. Therefore, this paper proposes a conversion from QNN that can handle the weights of various bit numbers.

\begin{figure}[]
    \centering
    \includegraphics[width=0.45\linewidth]{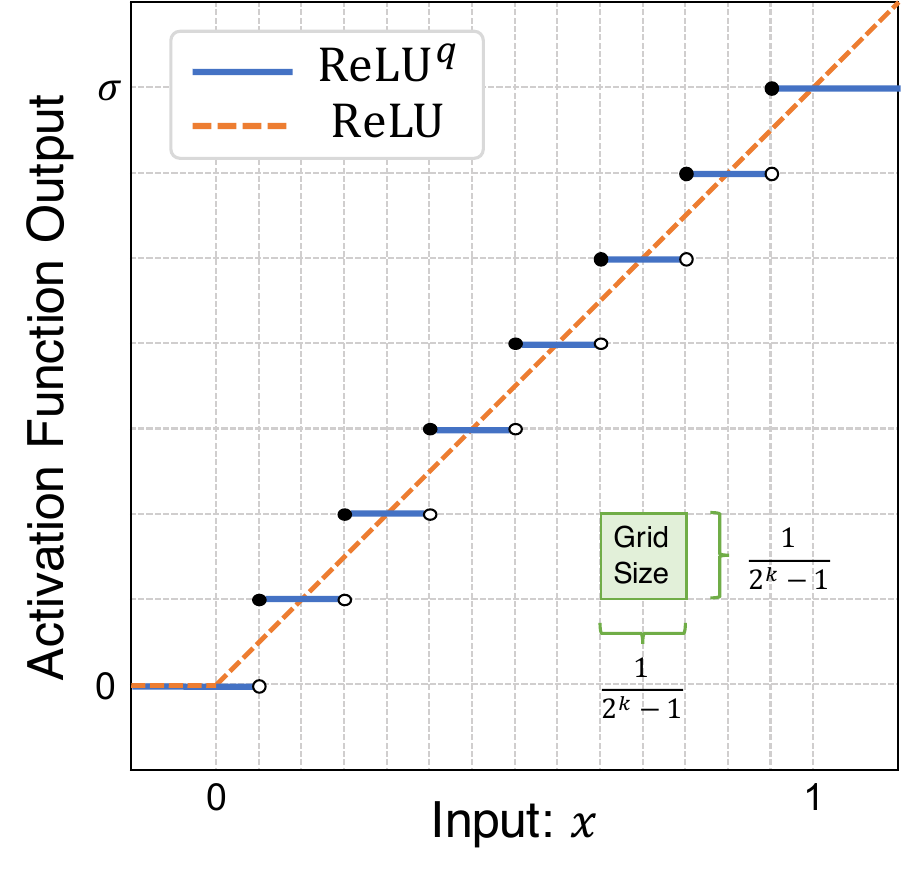}
    \caption{
        Difference between ReLU and $\text{ReLU}^q$:
        $k$ denotes quantization bit number.
        $\sigma$ denotes output scaling factor.
        ``Grid Size'' indicates quantization interval.
    }
    \label{fig:ReLU_compare}
\end{figure}

        This paper modifies the ReLU function since QNN can only represent values within the quantization-limited range, as in \equref{eq:quantize_activate_func}.
        QNN requires an activation function (shown in a previous work \cite{ANNtoSNN} and in \equref{eq:quantize_activate_func}), which cannot utilize the same conversion method as FPNN.
        Therefore, we propose an activation function that has the properties of both \equref{eq:ReLU} and \equref{eq:quantize_activate_func}: 
        \begin{align}
            \label{eq:ReLUforSNN}
            {\rm ReLU}^{q}(x) = \frac{\sigma}{2^k - 1}{\rm round} \left( \left( 2^k - 1 \right) \max(0, \min(1,x)) \right).
        \end{align}
        The plot of $\text{ReLU}^q$ is shown in \figref{fig:ReLU_compare}.
        A typical ReLU cannot be expressed by quantization bits because it takes up an infinite range. This paper limits ReLU's range with \equref{eq:ReLUforSNN} for quantizing it. The quantization range of each layer is fixed, and ReLU's output range is equally divided to set the quantization values. $\sigma$ is the maximum value of the quantization range, which controls the output scale of each node.

    \subsection{Robustification of Conversion Errors by Gap Increasing}
        To make the SNN policies robust to conversion errors, this paper adapts GIO to the QNN update procedure to emphasize the action order of the QNN policies.
        This paper expects that the GIO will keep the order robust to the conversion errors by QNN updates with GIO.
        The value function's gradient is calculated by $\bs{\theta}^{q}$ from learning samples $\mathcal{D} \ni (s, a, s', r^{a}_{s,s'})$.
        The loss function is estimated:
        \begin{equation}
            \begin{split}
                \label{eq:qpn-error} 
                & \hspace{-0.7em} J_{f}(s,a,s',r^{a}_{s,s'};\bs{\theta}^q,\bs{\theta}^{-q}) = 
                \frac{1}{2} \left[ r^{a}_{s,s'} + \gamma (m_{\beta}P)(s';\bs{\theta}^{-q}) \right . \\
                & \hspace{-0.7em} \left . +\alpha\left(P(s,a;\bs{\theta}^{-q}) \!-\! (m_{\beta} P)(s;\bs{\theta}^{-q})\right) - P(s,a;\bs{\theta}^q)) \right]^2,
            \end{split}
        \end{equation}
        where the $\alpha$ term denotes the GIO coefficient.
        Then $\bs{\theta}^{f}$ is updated by the gradient descent method for accurate gradient approximation.
        Noise is added to the gradient when estimating the loss by $\bs{\theta}^{q}$ for stable weight updates \cite{dorefa}.

\section{Experiments}
\label{experimens}
    This section evaluates the effectiveness of RIVC, our proposed method.
    We analyzed its features in simulation tasks of CartPole and a visual servo and demonstrated it in a neurochip-driven robot in a visual-servo task.

\subsection{Construction of Learning System for Experiments}
\label{settings-experimens}
    \subsubsection{Entire Experiment Settings}
        This section describes the construction of the proposed framework shown in \figref{fig:learning_system}.
        We utilized a desktop PC equipped with a GPU (Nvidia RTX3090) for updating the policies and an Akida Neural Processor SoC as a neurochip \cite{akida1,akida2}.
        The robot was controlled by the policies implemented in the neurochip.
        SNNs were implemented to the neurochip by a conversion executed by the MetaTF of Akida that converts the software \cite{akida1,akida2}.
        Samples were collected by the SNN policies in both the simulation tasks and the real-robot tasks since the target task is neurochip-driven robot control.
        For learning, the GPU updates the policies based on the collected samples in the real-robot environment.
        Concerning the SNN structure, the quantization of weights $w^s$ described in \equref{eq:snn_node} and the calculation accuracy of the activation functions described in \equref{eq:snn_activation} are verified in a range from 2- to 8-bits; they are the implementation constraints of the neurochip \cite{akida1}.

    \subsubsection{Previous Methods for Comparison}
        We compared the following methods to verify the effectiveness of the proposed method:
        \begin{itemize}
            \item DRL2SNN: Originally, this method optimized FPNN policies and converted trained FPNN policies to SNN policies. We evaluated it in experiments that modified it to train the FPNN policies and sequentially converted them to SNN policies after the former were updated.
            This method was evaluated as the related works of RIVC without a countermeasure of conversion errors.
            \item R-STDP: This method, which directly trains the SNN weights (excluding the SNN conversions), 
            was evaluated as a representative of the methods without BPs.
        \end{itemize}

        We also verified the following two ablation methods of RIVC:
        \begin{itemize}
            \item RIVC w/o GIO trains the QNN policies and sequentially converts them to SNN policies when the former are updated. It updates QNN without applying GIO.
            \item RIVC w/o Quantize trains the FPNN policies and sequentially converts them to SNN policies when the QNN policies are updated. It updates QNN by applying GIO.
        \end{itemize}

        We verified the following method to evaluate the performance's upper bound from methods using SNN conversion.
        \begin{itemize}
            \item FPNN-CVI: Conservative Value Iteration (CVI) \cite{cvi} trains FPNN policies, including learning by GIO, excluding SNN conversions.
            FPNN-CVI is utilized for the upper bound since CVI without this conversion does not degrade the learning performance.
        \end{itemize}

\begin{figure}[t]
    \centering
        \includegraphics[width=1\linewidth]{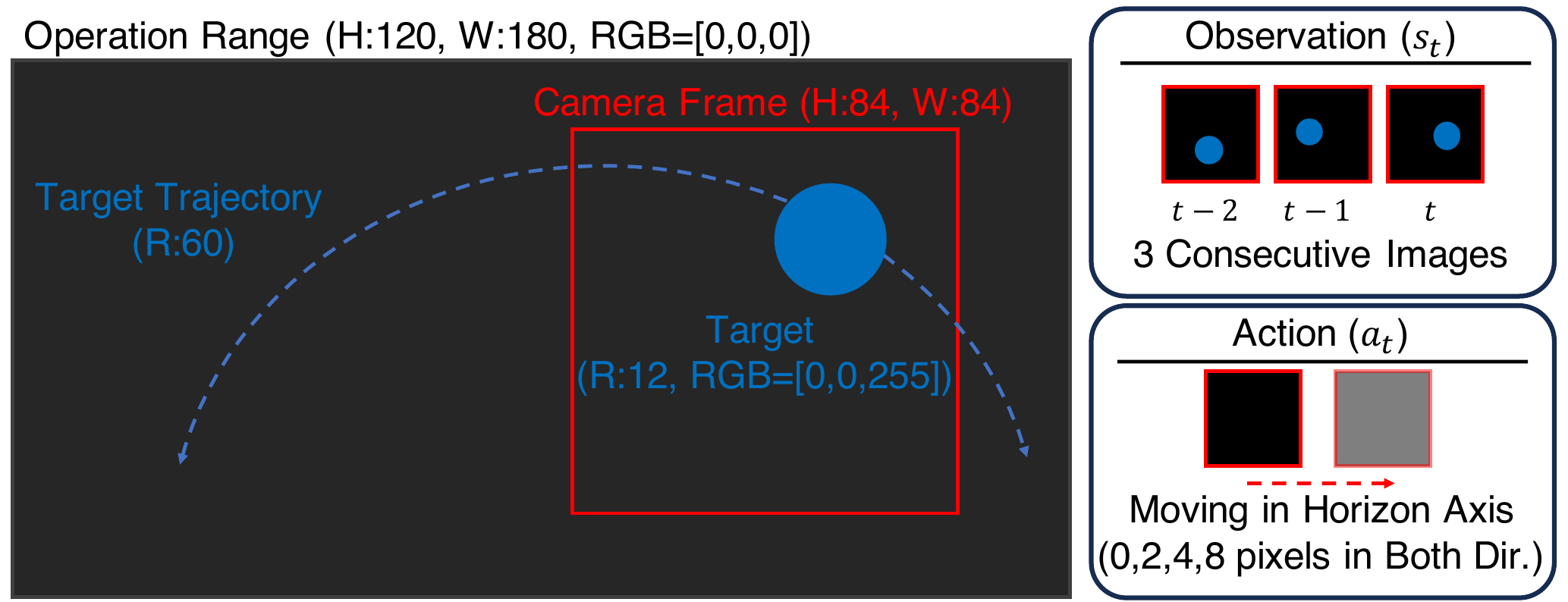}
        \caption{
             Simulation task settings of visual-servo task:
             Task environment is comprised of a ball (target object), a camera frame (agent), and a task field.
             Task objective is to track the ball by the camera frame.
             Agent's action is moving on horizon axis.
             Observation is three consecutive images in camera frames obtained by conversion from RGB to grayscale images.
             H, W, and R denote height, width, and radius expressed by pixels.
             RGB denotes RGB color values from 0 to 255.
        }
        \label{sim-setting-servo}
\end{figure}

\subsection{Simulation Experiments}
\label{simulation-experimens}

    In this section, we validate the performance of our proposed RIVC in a simulation and
    investigate the following:
    \begin{enumerate}
        \item Comparison of our proposed RIVC with previous works and ablation methods (Section \ref{sim:compare});
        \item Effect of the number of quantization bits of value functions on the learning performance of the total rewards, the stability, and the convergence (Section \ref{sim:qnn});
        \item Effect of the gap-increasing operator of the value updates for emphasizing the maximum action in RIVC (Section \ref{sim:gio}).
    \end{enumerate}
    
    \subsubsection{Settings}
        This experiment utilized two simulation environments: CartPole \cite{OpenAIgym} for a simple, low-dimensional vector observation task and a visual-servo task for a high-difficulty assignment of image observation shown in \figref{sim-setting-servo}.
        The details of each task are described as follows.
        
        \textbf{CartPole:}
            \label{sim:cartpole}
            In this experiment, we utilized OpenAIgym's CartPole \cite{OpenAIgym}.
            This task controls a cart to maintain a pole's balance.
            The action moves the cart left or right; thus the action dimension is $|\mathcal{A}|=2$.
            The observation is the cart position, its velocity, the pole angle, and the pole angular velocity; thus, the state dimension is $|\mathcal{S}|=4$.
            The network structure consists of four fully connected layers: FC(4), FC(256), FC(256), and FC(2).
            FC() denotes the fully connected layer; the parameter is the number of nodes.
            Each parameter is shown in \tabref{table:learning_setting}.
        
        \textbf{Visual Servo:}
            \label{sim:tracking}
            In this experiment, we utilized our original simulator of the visual-servo environment (\figref{sim-setting-servo}).
            This task's objective was to control the camera frame's movement so that the agent always captures the target within the camera frame.
            The action moves the frame left or right in 1, 2, or 4 pixels, including a stop action of 0 pixels; the action dimension is $|\mathcal{A}|=7$.
            The observation is three consecutive grayscale images from the current step to the last two steps for estimating the target's velocity and acceleration from an image series \cite{dqn}; the state dimension is $|\mathcal{S}|=84\times84\times3$.
            Reward $r$ is the Euclidean distance calculated by the center of camera frame $C^{\text{agent}}$ and the center of target $C^{\text{target}}$ as $r=-|C^{\text{agent}}-C^{\text{target}}|$.
            The initial positions of the agent and the target are fixed.
            The network structure consists of four convolution layers and three fully connected layers: Conv(3,16,7,2), Conv(16,32,5,2), Conv(32,64,5,1), Conv(64,64,3,1), FC(256), FC(256), FC(7).
            Conv() denotes a convolution layer, and the parameters are the numbers of input channels, output channels, kernels, and strides.
            Each parameter is identical, as in \tabref{table:learning_setting}, where $T=50$.

    \subsubsection{Comparison with Previous Works}
        \label{sim:compare}
\begin{figure}[t]
    \centering
    \begin{minipage}{0.49\linewidth}
        \centering
        \subfigure[CartPole]{
        \includegraphics[keepaspectratio, width=\linewidth, angle=0]
                    {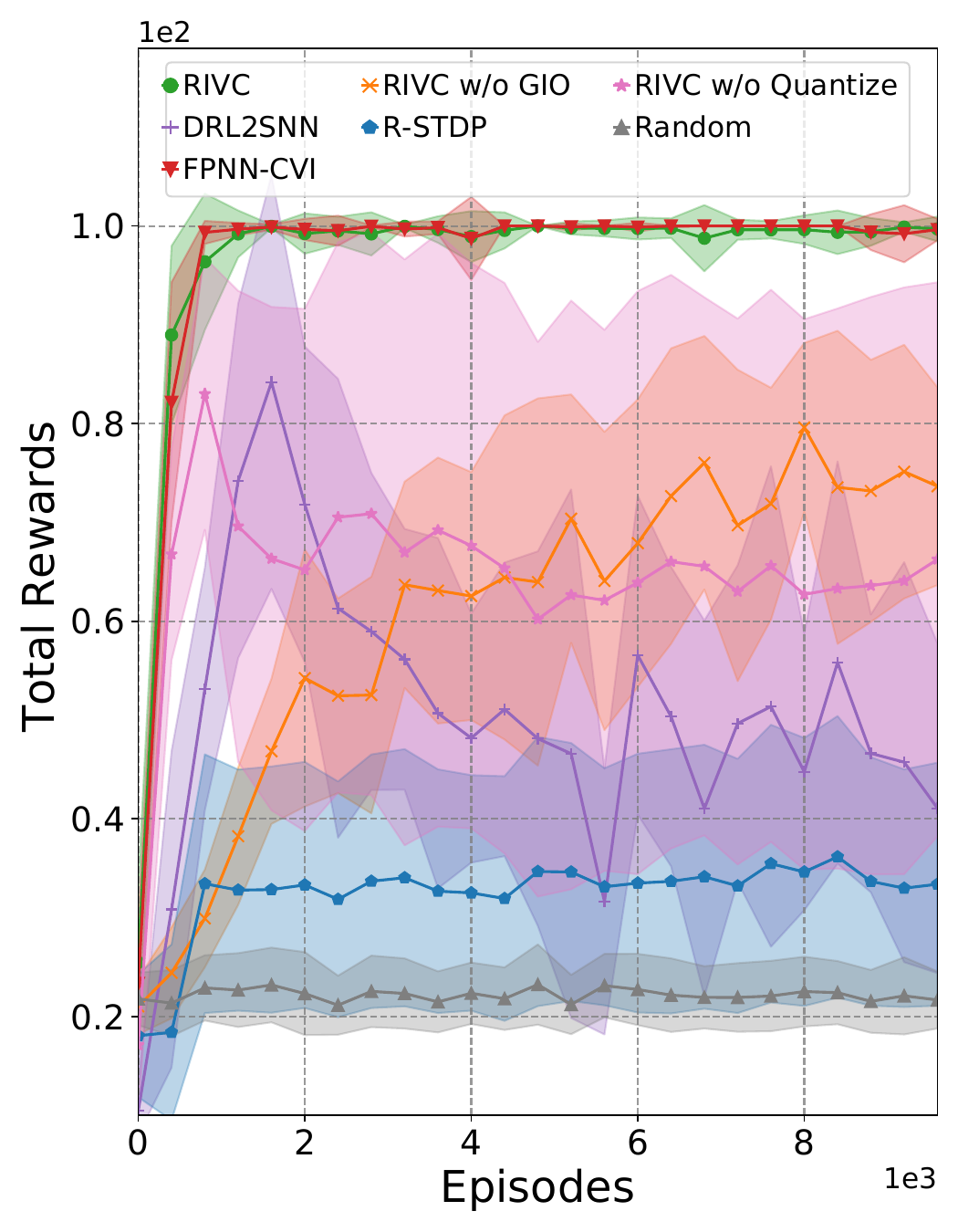}
        }
        \label{fig:sim_cartpole_all_learning_method}
    \end{minipage}
    % \vspace{5mm}
    \begin{minipage}{0.49\linewidth}
        \centering
        \subfigure[Visual Servo]{
        \includegraphics[keepaspectratio, width=\linewidth, angle=0]
                    {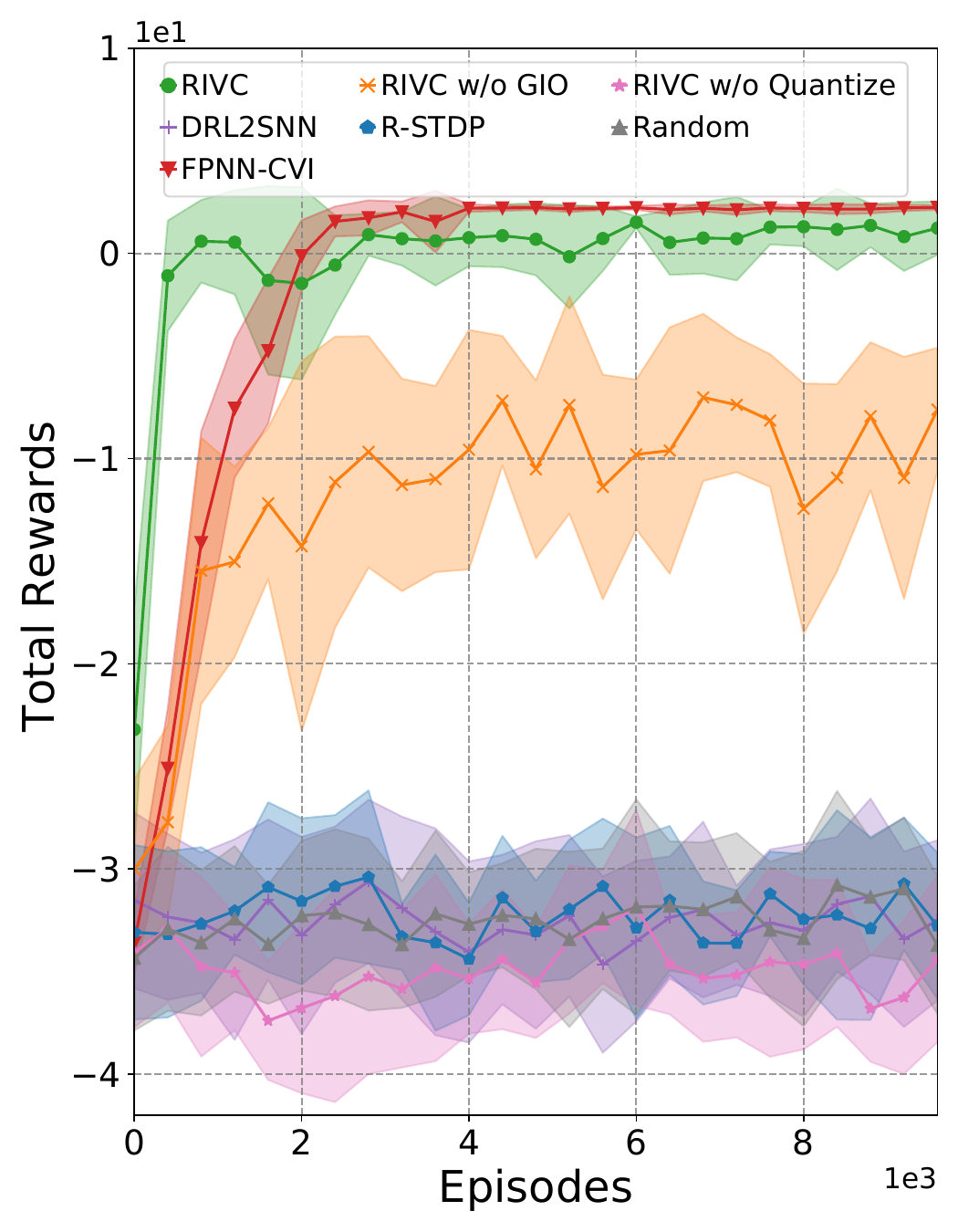}
        }
        \label{fig:sim_tracking_all_learning_method}
    \end{minipage}
    \caption{
        Comparison of learning methods: a) CartPole and b) Visual Servo. 
        Four-bit quantization is applied to RIVC and DRL2SNN.
        Each figure curve plots mean and variance per sample over five experiments.
    }
    \label{fig:sim_compare_all}
\end{figure}

        To compare the performance of the proposed RIVC with other methods, this experiment evaluated the RIVC ablation methods, DRL2SNN, R-STDP, and FPNN-CVI (\figref{fig:sim_compare_all}).

        In both tasks, RIVC achieved the highest total rewards within the SNN policy training methods.
        RIVC w/o GIO and RIVC w/o Quantize achieved approximately $75 \%$ of RIVC in the CartPole task, which was easy due to the low-dimension observations.
        The performance gap between RIVC and the other ablation methods was caused by the conversion errors.
        On the other hand, only RIVC w/o GIO achieved approximately $83 \%$ of RIVC in the visual-servo task, which was difficult due to the high-dimensional image observations.
        R-STDP only achieved approximately $14 \%$ of RIVC in the CartPole task and didn't progress at all in the visual-servo task.
        
        Based on these results, RIVC w/o GIO did not outperform RIVC.
        Nor did DRL2SNN. It obtained the SNN policies by converting the FPNN policies, which obtained the SNN policies by converting the QNN policies.
        R-STDP cannot train the policies with sufficient performance.
        As a result, only the proposed RIVC solved the tasks due to its treatment of robust conversion errors.
        Such robustness to conversion errors is verified in detail in \chapref{sim:gio}.

        Compared to the upper bound FPNN-CVI, RIVC achieved approximately an equivalent performance. Although RIVC improved the reward faster in the early learning stages, FPNN-CVI was better at converging to the best performance. This phenomenon is discussed in \chapref{discussion}.

\begin{figure}[]
    \vspace{2mm}
    \centering
    \begin{minipage}{0.49\linewidth}
        \centering
        \subfigure[CartPole]{
        \includegraphics[keepaspectratio, width=\linewidth, angle=0]
                    {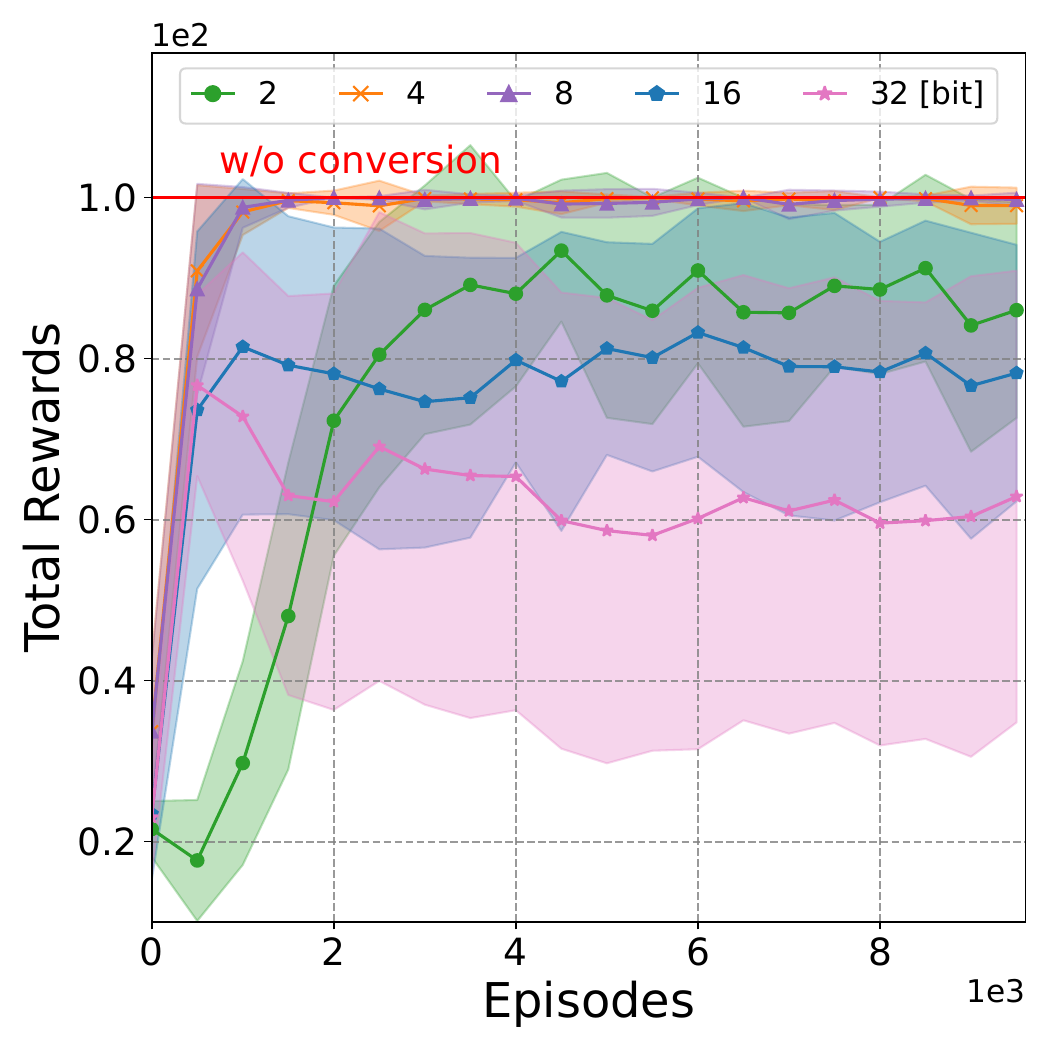}
        }
        \label{fig:sim_cartpole_quantized}
    \end{minipage}
    \begin{minipage}{0.49\linewidth}
        \centering
        \subfigure[Visual Servo]{
        \includegraphics[keepaspectratio, width=\linewidth, angle=0]
                    {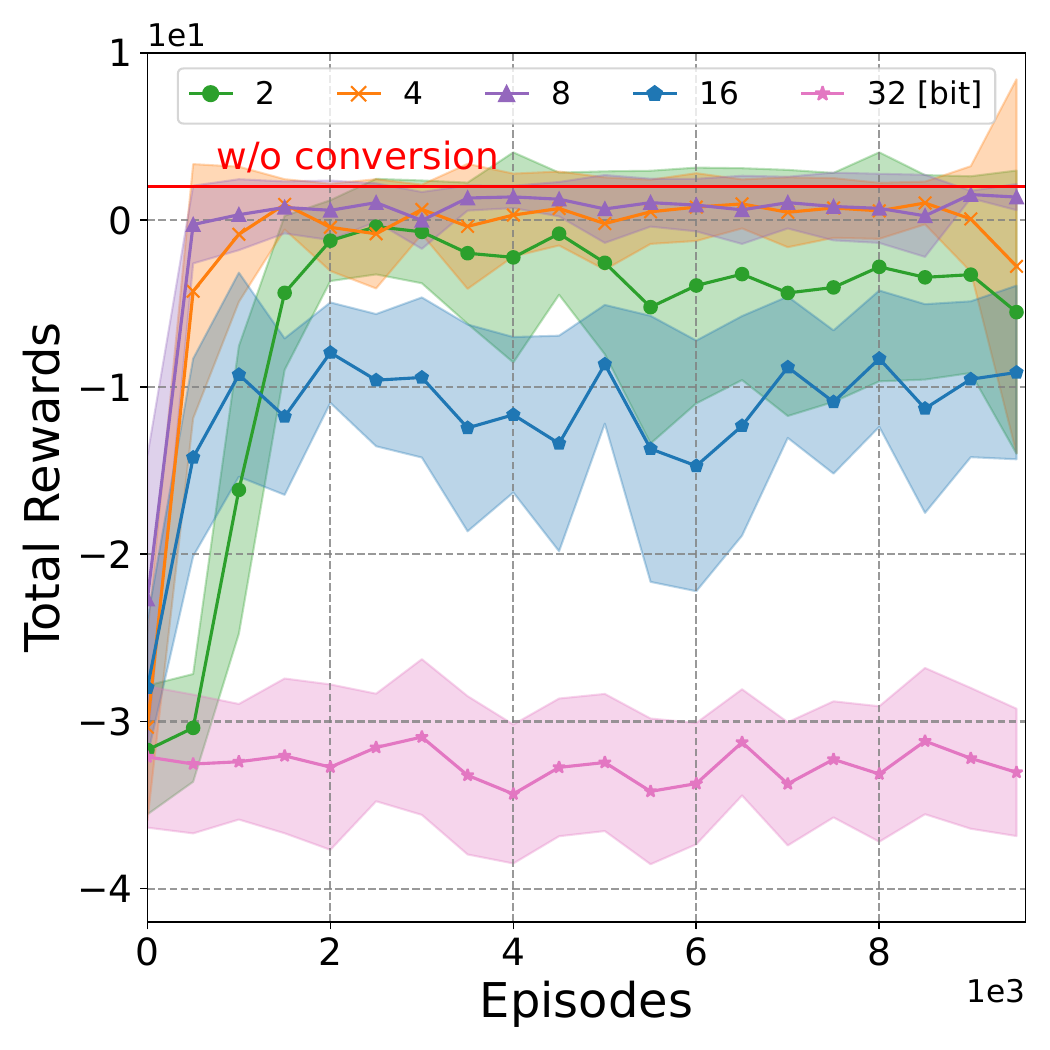}
        }
        \label{fig:sim_tracking_quantized}
    \end{minipage}
    \caption{
        Comparison of RIVC learning curves for each amount of quantization bits: a) CartPole and b) Visual Servo.
        Each plot number shows quantization bit of weights.
        Graph numbers denote quantization bits of weights and calculation accuracy of activation functions.
        ``w/o conversion'' lines denote maximum total reward achieved by FPNN-CVI which trains FPNN policies without conversion to SNN. 
        Each figure curve plots mean and variance per sample over five experiments.
    }
    \label{fig:sim_quantize_bit}
\end{figure}

    \subsubsection{Evaluation of Best Quantization-bit number from Learning Performance}
        \label{sim:qnn}
        This experiment evaluated the effect of the number of quantization bits of the QNN on learning performances.
        We evaluated whether the learning performance is high when the quantization bits are close to those of SNNs. This experiment was inspired by previous work \cite{fnn2bnn-snn-1,fnn2bnn-snn-2,fnn2bnn-snn-3} that utilized conversion from binarized neural networks (BNNs) with 1-bit weights and 1-bit activation. 
        
        This experiment evaluated the number of bits with the best learning performance, such that the quantization error was small and the approximation accuracy was not considerably reduced by quantization.
        Thus, we compared the learning performance by various quantization bits.
        The results (\figref{fig:sim_quantize_bit})
        show that the highest performance (in total reward and convergence speed) was obtained in 4- or 8-bits, followed by 2-, 16-, and 32-bits.
        These results confirm that the higher the NN quantization weights, the lower is the performance of the converted SNN.
        
        On the other hand, lower weights are not necessarily better; around 4- or 8-bits are appropriate.
        The most significant factor causing this result is that 8 is the upper limit of the number of quantization bits of the neurochip.
        Since the trained policies need to convert to the same bit as the SNN policies, the quantization of the policy conversion results in large changes in the parameters of the policies trained with weights over 8 bits.
        Therefore, the performances of the 16- and 32-bit policies were significantly degraded.
        The latter's performance was even worse than the former's due to the large difference in the number of quantization bits between the 32-bit policies and the SNN policies.

        The other experiments used 4-bit quantization for training QNNs based on two aspects: 1) these experiments show that 4- or 8-bits achieved the best performance, and 2) a QNN with lower bits can be calculated quickly.

    \subsubsection{Evaluation of GIO Effect for Policy Action Order}
        \label{sim:gio}
\begin{figure}[]
    \vspace{2mm}
    \centering
    \begin{minipage}{0.49\linewidth}
        \centering
        \subfigure[CartPole]{
        \includegraphics[keepaspectratio, width=\linewidth, angle=0]{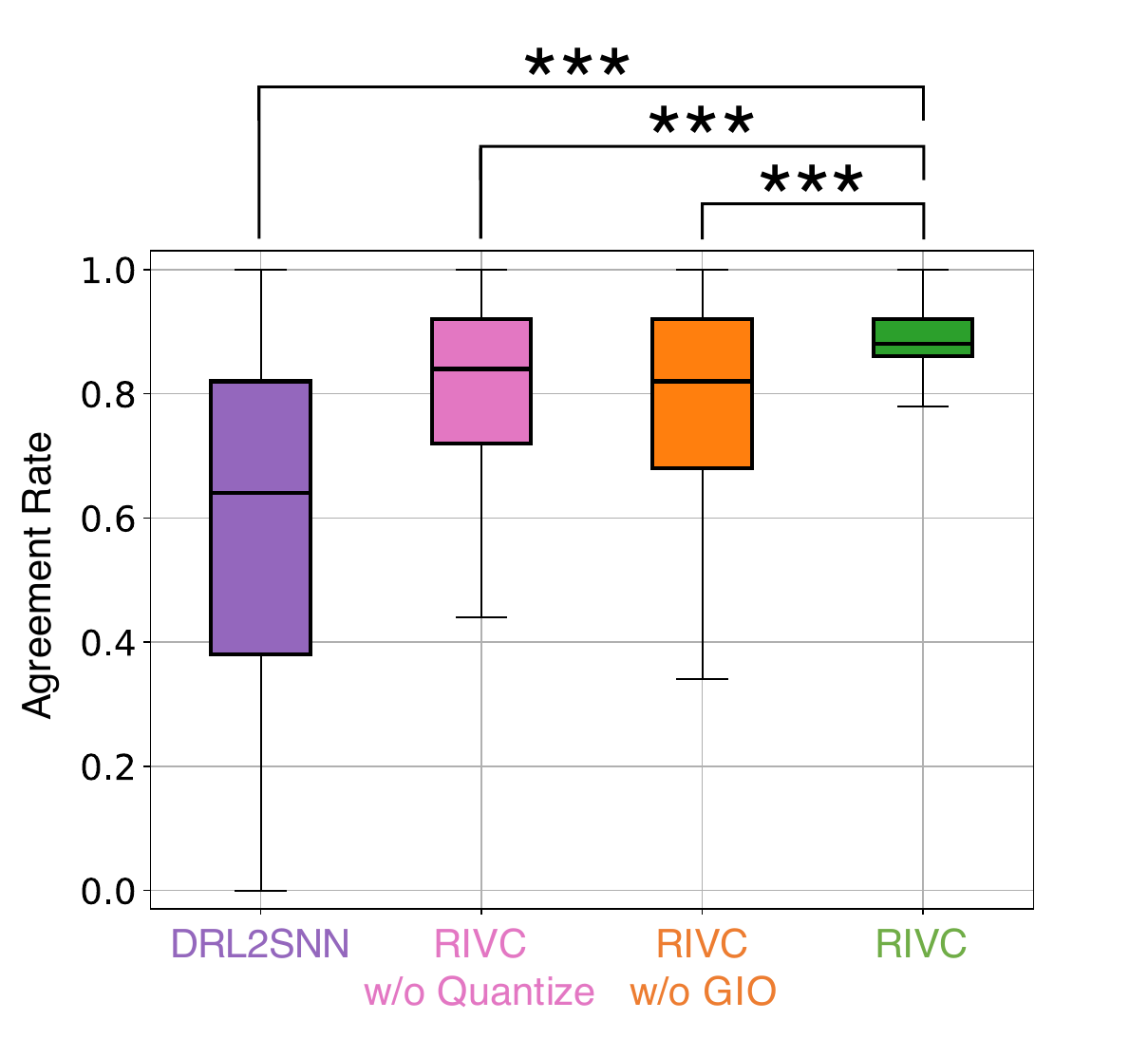}
        }
        \label{fig:sim_tracking_action_index_loss_c}
    \end{minipage}
    \begin{minipage}{0.49\linewidth}
        \centering
        \subfigure[Visual Servo]{
        \includegraphics[keepaspectratio, width=\linewidth, angle=0]{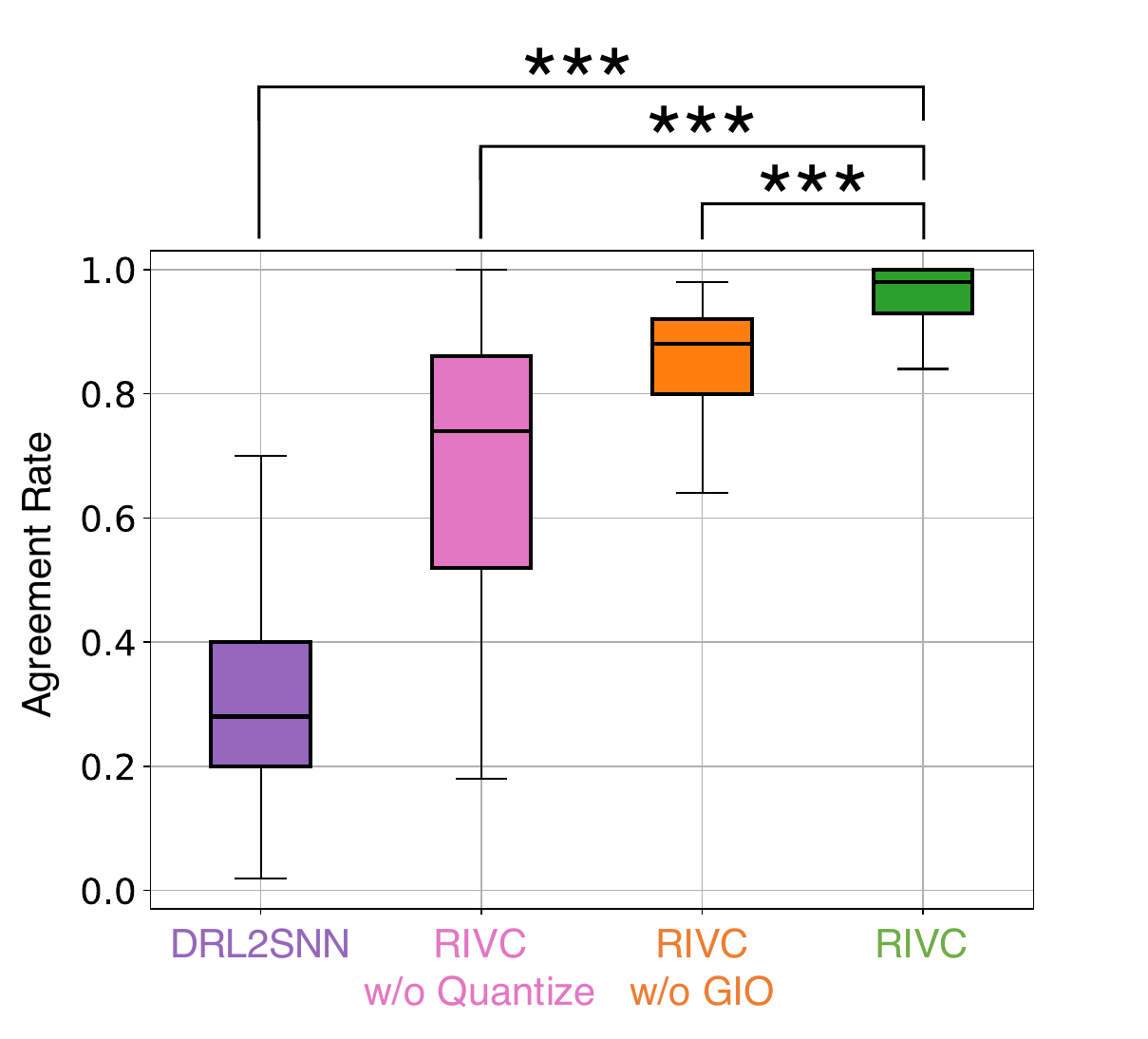}
        }
        \label{fig:sim_tracking_action_index_loss_v}
    \end{minipage}
    \caption{
        Agreement rate of actions corresponding to maximum value $P$ of conversion between QNN and SNN:
        Agreement rate is defined as $\sum_{s \in \mathcal{D}} \delta \left( \argmax_a(P(s,a;\theta^{q})), \argmax_a(P(s,a;\theta^{s})) \right)$, where $\delta$ means a Kronecker delta function.
        Each boxplot evaluated entire step agreement rate per experiment (\textasteriskcentered\textasteriskcentered\textasteriskcentered \ means $p <  0.001$).
        Each boxplot was evaluated over five experiments.
        Quantization bits of weights and calculation accuracy of activation functions are 4-bits.
        These experiments were evaluated in visual-servo tasks.
    }
    \label{fig:sim_tracking_action_index_loss}
\end{figure}

        This experiment verified GIO's effect, which
        emphasizes the value gap between the probabilities of selecting an optimal action and other actions.
        We expected that a GIO can prevent the alternation of the maximum actions caused by SNN conversion.
        Thus, we evaluated the agreement rates of the policy's maximum actions before and after conversion in settings with/without GIO.
        The results are shown in \figref{fig:sim_tracking_action_index_loss}.
        
        RIVC outperformed the other methods for every quantization pattern regarding the agreement rates, particularly its approximately over 80\% agreement rate for both tasks.
        On the other hand, DRL2SNN showed a large variation in its agreement rate, reaching 0\% at its low values. The average agreement rate was also less than 60\%.
        For the other ablation methods, the agreement rate exceeded 70\% on average, although the variance was also large. 
        For the visual-servo task in \figref{fig:sim_compare_all}, only RIVC accomplished it. However, RIVC w/o GIO achieved 80\% of RIVC's performance because of its low variance, and its agreement rates exceeded 80\% on average.

\subsection{Real-robot Experiments}
\label{real-experiments}

    In this section, we verify the effectiveness of the proposed RIVC on a real-robot task (\figref{fig:overview}):
    an object-tracking task in a real-robot environment. The goal is to track a moving ball in real-time with a camera robot.
    This is an appropriate real-time image-input task because it requires real-time responses where the neurochip needs to calculate the image-input SNN policies to the output actions, including both communication and robot control delays.
    
    \subsubsection{Settings}
\begin{figure}[t]
    \centering
    \includegraphics[width=0.9\linewidth]{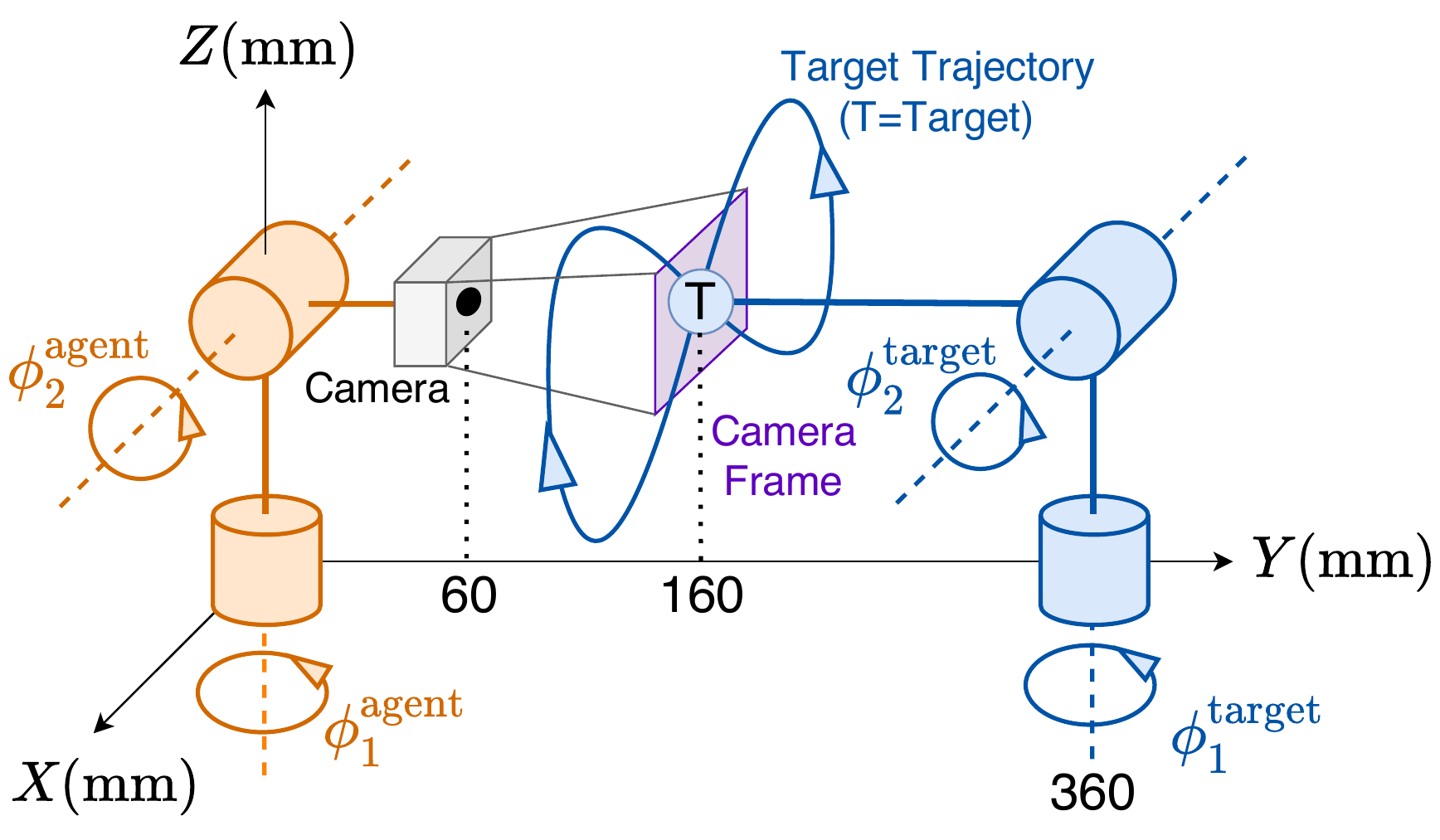}
    \caption{
        Setting up environment for real-robot object-tracking task:
        Learning environment is comprised of agent (orange) and target controller (blue).
        Agent controls two motors $(\phi_{1}^{\text{agent}},\phi_{2}^{\text{agent}})$ to track target within camera frame (purple).
        Target controller operates two motors $(\phi_{1}^{\text{target}}, \phi_{2}^{\text{target}})$ to manipulate target in a figure-8 pattern.
    }
    \label{fig:visual_servo_task_design}
\end{figure}

        An overall view of the learning environment is shown in \figref{fig:overview}, and the experimental setup's details are shown in \figref{fig:visual_servo_task_design}.
       This task's objective is to control the robot so that it keeps the ball in the camera frame.
        The agent and target each consist of two servo motors (Dynamixel XM430-W350-T) manipulated by position control.
        Observation $s$ is an $84 \times 84$ pixel grayscale image input for three consecutive steps.
        Action $a$ controls two motors as eight cardinal directions with 4-degree rotation. The action dimension is $|\mathcal{A}|=8+1$, including stop action \cite{dqn}.
        The definitions of reward and episode are identical as in \chapref{sim:tracking}.
        The learning parameters and the network structure are identical as in \tabref{table:learning_setting}, except that $I=30$, $T=30$, and the QNN weights and the calculation accuracy of the activation functions are 4-bits.

\begin{figure}[]
        \hspace{-2mm}
        \begin{minipage}[b]{0.4\linewidth}
            \centering
            \vspace{-2mm}
            \subfigure[Learning Curves of figure-8]{
            \label{fig:real_result_learning_curve}
            \includegraphics[keepaspectratio, width=\linewidth, angle=0]
                        {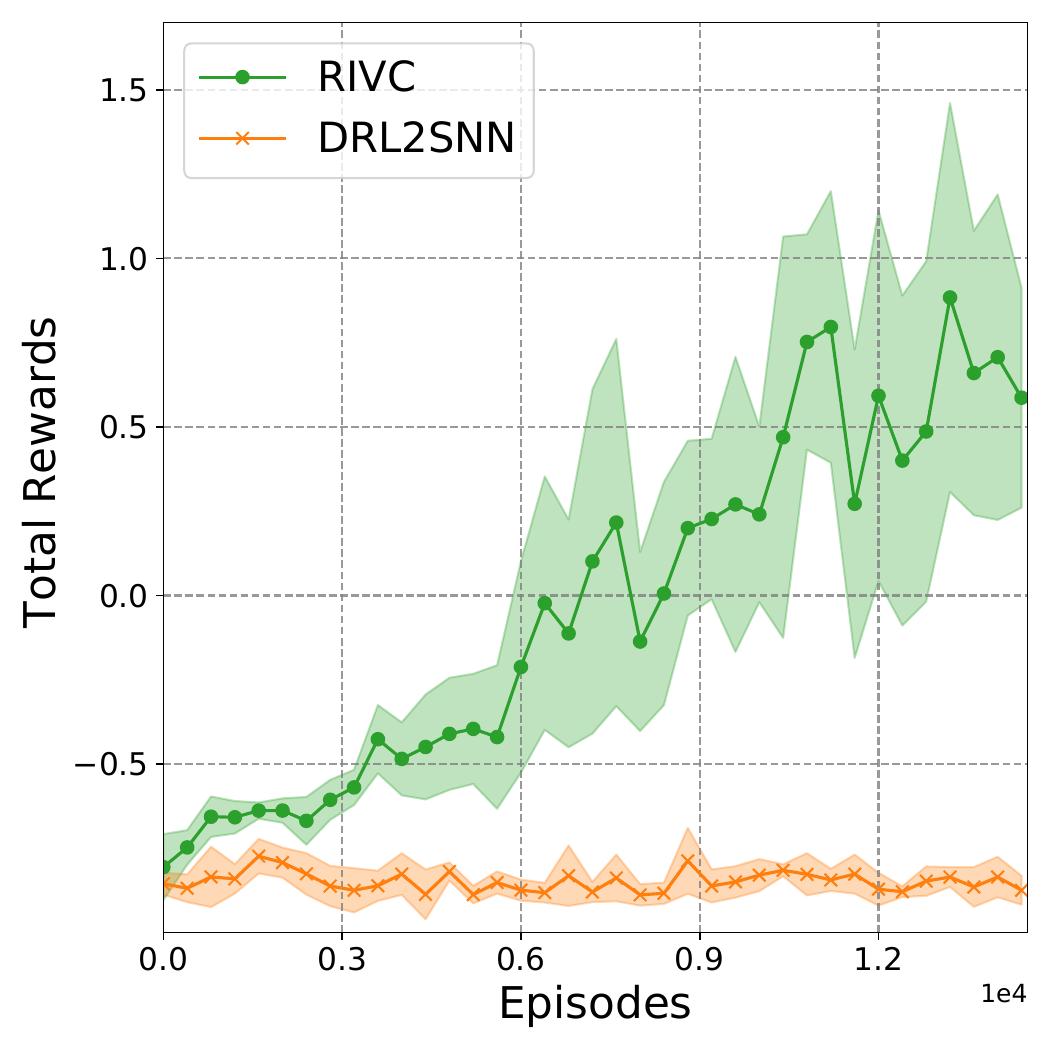}}
        \end{minipage}
        \hspace{-1mm}
        \begin{minipage}[b]{0.42\linewidth}
            \centering
            \subfigure[Trajectories of figure-8]{
            \label{fig:real_result_trajectories}
            \includegraphics[keepaspectratio, width=\linewidth, angle=0]
                        {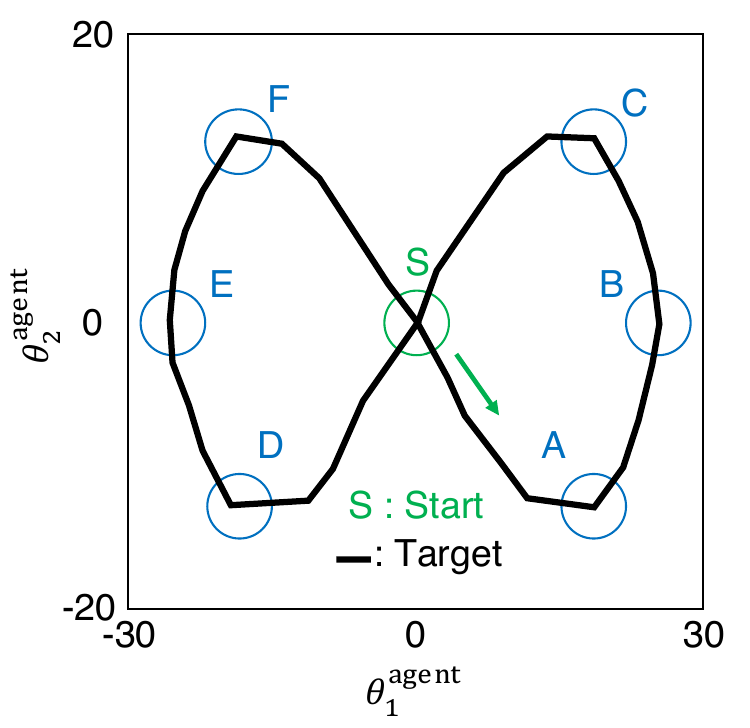}}
        \end{minipage}
        \\
        \begin{minipage}[b]{0.4\linewidth}
            \centering
            \vspace{-2mm}
            \subfigure[Learning Curves of Random]{
            \label{fig:real_result2_learning_curve}
            \includegraphics[keepaspectratio, width=\linewidth, angle=0]
                        {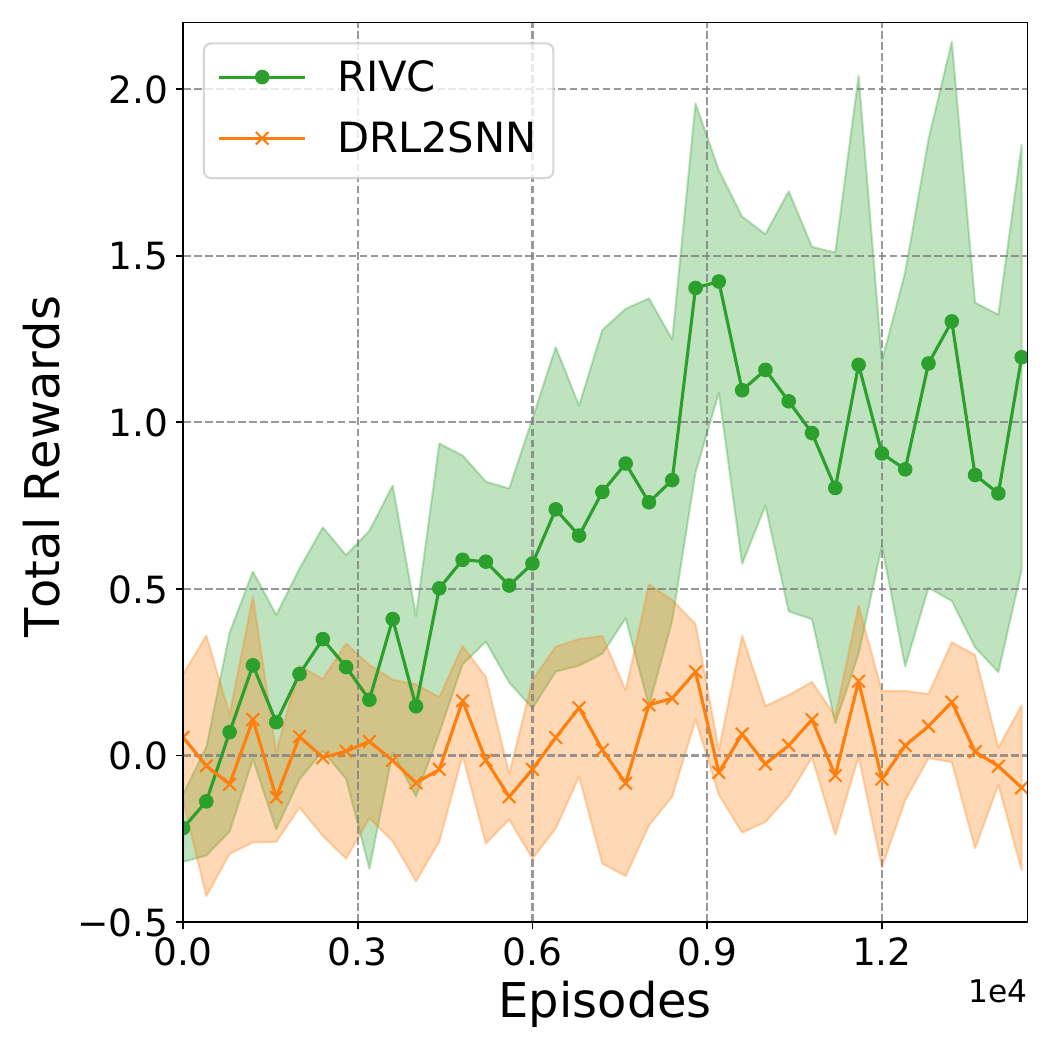}}
        \end{minipage}
        \hspace{-1mm}
        \begin{minipage}[b]{0.42\linewidth}
            \centering
            \subfigure[Trajectories of Random]{
            \label{fig:real_result2_trajectories}
            \includegraphics[keepaspectratio, width=\linewidth, angle=0]
                        {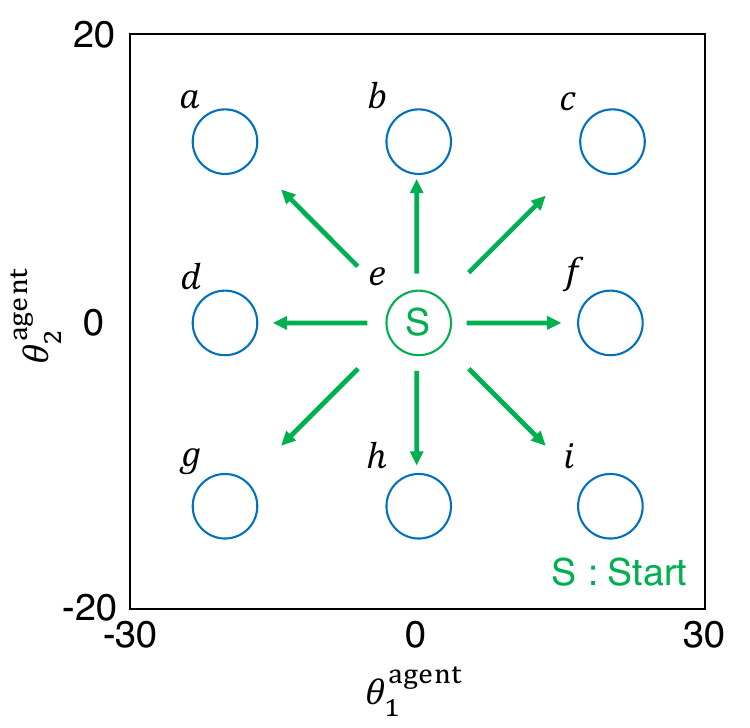}}
        \end{minipage}
        \\
        \hspace{-2mm}
        \begin{minipage}{0.9\linewidth}
            \centering
            \subfigure[Observations]{
            \label{fig:real_result_observations}
            \includegraphics[keepaspectratio, width=\linewidth, angle=0]
                        {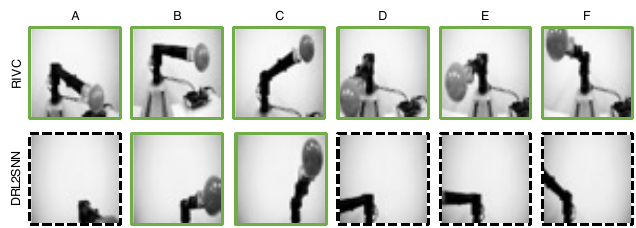}}
        \end{minipage}
    \caption{
        Learning results of real-robot visual-servo task:
        Learning curves plot mean and variance of total reward per iteration $I$ over five experiments.
        Target trajectories of two motors of agent, $\bs{\theta}_{1}^{\text{agent}}$ and $\bs{\theta}_{2}^{\text{agent}}$, are figure-8 patterns in (\textbf{b}) and random patterns in (\textbf{d}).
        (\textbf{e}) shows observation from camera at points of figure-8 plotted in (\textbf{b}) as A to F.
        Observations are $84 \times 84 $ pixels.
        As of (\textbf{e}), green line frame and black dashed-line frame indicate tracking success and failure, respectively.
        Quantization-bit number is 4 for evaluating RIVC and DRL2SNN.
    }
    \label{fig:real_result-all}
\end{figure}

        In this experiment, we learned two types of ball-trajectory tracking.
        \begin{itemize}
            \item figure-8 trajectory of \figref{fig:real_result_trajectories}: The trajectories of the target's motor angle $(\phi_{1}^{\text{target}},\phi_{2}^{\text{target}})$ were calculated with angular velocity $\omega$ and time step $t$ as $\phi_{1}^{\text{target}} = 25 \sin{\omega t}$, $\phi_{2}^{\text{target}} = 15 \sin{2 \omega t}$, $\omega = \pi/5$ rad/s.
            \item random trajectory of \figref{fig:real_result2_trajectories}: The ball moves randomly in the nine points, with horizontal-point spacing at 25 degrees and vertical-point spacing at 15 degrees.
            After the ball arrives at each point, it repeatedly moves to a randomly chosen point.
        \end{itemize}

    \subsubsection{Learning Control Policies}
        The learning results are shown in \figref{fig:real_result-all}.
        From \figref{fig:real_result_learning_curve} and \figref{fig:real_result2_learning_curve}, DRL2SNN failed to learn due to conversion errors, although RIVC successfully trained the policies. DRL2SNN cannot consistently capture the ball in the camera frame, although RIVC can.

        \figref{fig:real_result_observations} is an observation of the trained policy when the target is in the agent's target trajectory A to F in \figref{fig:real_result_trajectories}.
        RIVC can track the target to remain in the camera frame using raw image inputs containing wiring, a chip, and robots.
        However, DRL2SNN is out of the frame from point A.
        The target returns coincidentally to the camera frame at point B, but DRL2SNN is again out of the frame at point D.
        
        Compared to the simulation visual-servo task that only controlled the camera frame on the horizon axis, this real-robot visual-servo task needs to control the camera frame on the two-axis and include some environmental noise (e.g., shadows from the light conditions and objects other than balls).
        When the task becomes noisier, RIVC can train the policies by increasing the value gaps between the optimal actions and others, but DRL2SNN fails for a lack of such structures.

    \subsubsection{Comparison of Calculation Speed and Power Consumption}
        We next verified whether the neurochip can calculate the SNN policies in real-time and with low power consumption.
         Based on calculation speed and power consumption, we first compared whether the calculated SNN policies in the neurochip outperformed the calculated FPNN policies in the edge CPU.
        These terms were evaluated by policy calculations within the duration from an observation's input to an action's output.
        The calculation settings are identical as the real-world visual-servo task.
        
        The results are shown in \tabref{table:consumed_energy}.
        The neurochip's power consumption is approximately 15 times less than the edge CPU, and its calculation speed is about five times faster.
        Thus, the neurochip is better for edge robots with limited battery capacity in real-time operations.
        Applicable tasks include real-time object avoidance and trajectory tracking for robots.

\begin{table}[t]
    \vspace{1.mm}
    \begin{center}
    \caption{
        Hardware performance of policies:
        FPNN was evaluated by edge-CPU (Raspberry Pi 4: quad-core ARM Cortex-A72).
        SNN was evaluated by neurochip (Akida 1000 \cite{akida1}).
        ``Power cons'' and ``Calc. speed'' denote power consumption and calculation speed for obtaining one action from NN policies using each piece of hardware.
        Power consumption was measured by voltage checker (TAP-TST8N).
        \label{table:consumed_energy}
    }
        \begin{tabular}{lccc}
            \toprule
            Hardware & Edge-CPU & Neurochip \\
            Network & FPNN & SNN \\
            \midrule
            Power consumption [mW] & $61$ & $4$ \\
            Calculation speed [ms] & $205$  & $40$ \\
            \bottomrule   
        \end{tabular}
    \end{center}
\end{table}

\section{Discussions}
\label{discussion}

    From \figref{fig:sim_tracking_action_index_loss}, we confirmed that GIO suppresses the changes of the actions of the maximum values.
    The difference in the agreement rate of the maximum action is twice that between RIVC and DRL2SNN; a noticeable difference is seen in the learning performance.
    For the visual-servo task, the difference in the agreement rate with/without GIO is approximately 10\%, but the minimum number of agreement rates without GIO is approximately 20\% lower than with GIO. 
    As shown in \figref{fig:sim_compare_all}, the difference in the learning performance of RIVC and RIVC w/o GIO is tremendous, indicating that the replacements in the maximum actions significantly impact the learning performance if the agreement rates are approximately 10\% different.
    Even just a slight percentage difference in the agreement rate can create many completely unintended actions in one episode. As a result, a task often fails due to outputting the wrong action, suggesting that RIVC w/o GIO performs poorly and RIVC performs well.

    From \figref{fig:sim_quantize_bit}, the optimal bit number of the weights is not the maximum or minimum of the available range of the quantization-bit numbers, although a certain optimal number does exist. In this case, RIVC's learning performance is best when the weights are 4- and 8-bit.
    When the weights are 16- and 32-bit cases, RIVC's learning performance is the worst.
    The learning performance steadily decreases when the weights exceed 16-bits, perhaps because the conversion error is bigger for large quantization.
    When the weights are 2-bits, RIVC's learning performance is better than for the 16- and 32-bit cases and worse than the 4- and 8-bit cases.
    This is because RIVC's learning performance is higher when the quantization bits are closer to those of SNNs, as shown by previous works \cite{fnn2bnn-snn-1,fnn2bnn-snn-2,fnn2bnn-snn-3} that utilized conversion from binarized neural networks (BNNs) with 1-bit weights and 1-bit activation.
    Converting from BNNs to SNNs indicates that learning with NNs performs better because NNs' quantization bits are close to those of the SNNs.

    Throughout the simulation tasks, we confirmed that the previous methods learned CartPole, which is a low-dimensional vector observation task.
    On the other hand, only the proposed method learned the visual-servo task, which is a high-dimensional image-observation task.
    There are two differences between the proposed RIVC and the other methods: 1) learning with quantized weights and 2) learning with an increasing gap.
    As shown in \figref{fig:sim_quantize_bit}, the performance on the image-input task is poor when converting to SNN policies from 32-bit FPNN policies.
    As shown in \figref{fig:sim_tracking_action_index_loss}, the replacement of maximum actions by SNN conversion is prevented by GIO. In addition, gap increases may be necessary for deciding the sub-optimal actions for exploration. 
    RIVC's two features are essential.

    \figref{fig:sim_tracking_action_index_loss} shows that even a RIVC robust to conversion error has 10-20\% mistakes when converting maximum actions.
    Our experiments show that learnings are successful since the policies can overcome such an amount of conversion errors. 
    On the other hand, conversion errors may accumulate in longer horizon tasks. 
    The total number of steps for CartPole is 100 and 50 for visual servo.
    Comparing the two tasks, CartPole, which has more total steps, has a lower agreement rate, which might fall
    due to the accumulative conversion errors when applied to long-horizon tasks.
    To solve this problem, future work must investigate further countermeasures against conversion errors.

    \figref{fig:sim_compare_all} shows that FPNN-CVI more quickly converged to the best performance while RIVC improved the reward faster in the early stages of learning. There are two possible reasons for this phenomenon.
    First, quantization restricts the values of the NN parameters, and previous research has shown that it acts as a regularization effect to prevent overfitting to the learning samples \cite{qnn_normalization_effect1,qnn_normalization_effect2}. Therefore, this regularization effect might stabilize the learning process and improve the policy performance quickly, even with a small amount of learning samples.
    Second, note that optimal parameters can be found efficiently since quantization discretizes the parameter space of the NN and narrows the range of parameters to be explored \cite{qnn_explore_effect1,qnn_explore_effect2}. This phenomenon might have accelerated learning.
    On the other hand, FPNN-CVI's final performance converged more quickly. Although QNNs have regularization effects, their function approximation performance is also limited due to the restricted parameter space. In reinforcement learning as learning progresses, the total rewards must be approximated for longer-term state-action transitions, requiring high function approximation performance. In this respect, FPNN has a higher function approximation performance than QNN, suggesting that it converges to high total rewards more quickly than QNN.

\section{Conclusion}
\label{conclusion}

    We proposed RIVC as a novel DRL framework for training SNN policies with a neurochip in real-robot environments.
    RIVC offers two prominent features: 1) it trains QNN policies, which can be robust for conversion to SNN policies, and 2) it updates the values with GIO, which is robust to the optimal action replacements by conversion to SNN policies.
    We also implemented RIVC for object-tracking tasks with a neurochip in real-robot environments.
    Our experiments show that RIVC can train SNN policies by DRL in real-robot environments.

\section*{Acknowledgments}
    This work was supported by the MegaChips Corporation.
    We thank Alonso Ramos Fernandez for his experimental assistance.

\bibliography{paper}

\end{document}